\newif\ifnonanonymous
\newif\ifuseappendix
\newif\ifuseacknowledgement
\newif\ifwacv

\nonanonymoustrue 
\useacknowledgementtrue 
\useappendixtrue 
\wacvtrue 

\ifwacv
\documentclass[10pt,twocolumn,letterpaper]{article}
\else
\documentclass{article}
\fi

\usepackage[T1]{fontenc}    

\ifnonanonymous
    \ifwacv
        \usepackage[pagenumbers]{wacv} 
    \else
        \usepackage[nonatbib,final]{neurips_2025}
    \fi
\else

    \ifwacv
        \usepackage[review,applications]{wacv}      
    \else
        \usepackage[nonatbib]{neurips_2025}
    \fi

\fi

\usepackage[table,xcdraw,dvipsnames]{xcolor}

\ifwacv
\definecolor{wacvblue}{rgb}{0.21,0.49,0.74}
\usepackage[pagebackref,breaklinks,colorlinks,allcolors=wacvblue]{hyperref}
\def\confName{WACV}
\def\wacvPaperID{3} 
\def\confName{WACV WasteVision}
\def\confYear{2026}
\else

\usepackage[
  breaklinks=true,
  pagebackref=true,
  colorlinks=true,      
]{hyperref}
\fi

\usepackage{graphicx}
\usepackage{amsmath}
\usepackage{amssymb}
\usepackage{framed}
\usepackage{booktabs}
\usepackage{comment}
\usepackage{makecell}
\usepackage{wrapfig}
\usepackage{subcaption}
\usepackage{cclicenses}
\usepackage{xspace}
\usepackage{pdfcomment} 

\ifwacv

\usepackage{caption}
\else
\usepackage{multicol}
\fi

\usepackage{amsfonts}       
\usepackage{nicefrac}       
\usepackage{microtype}      
\usepackage{lipsum}         
\usepackage[numbers]{natbib}
\usepackage{doi}
\usepackage{seqsplit}
\usepackage{placeins}

\usepackage{listings}
\usepackage[toc]{appendix}

\usepackage[capitalize]{cleveref}
\crefname{section}{Sec.}{Secs.}
\Crefname{section}{Section}{Sections}
\Crefname{table}{Table}{Tables}
\crefname{table}{Tab.}{Tabs.}

\ifwacv
\else
\newcommand{\eg}{e.g.\@\xspace}

\fi

\newcommand{\redact}[1]{%
    \ifnonanonymous
        #1
    \else
        [redacted for peer review]
    \fi
}

\newcommand{\cotwo}{\ensuremath{\mathrm{CO_2}}}

\newcommand{\ipfsgateway}{https://ipfs.io/ipfs/QmQonrckXZq37ZHDoRGN4xVBkqedvJRgYyzp2aBC5Ujpyp?autoadapt=0&requiresorigin=0&web3domain=0&immediatecontinue=1&magiclibraryconfirmation=0&redirectURL=}

\newcommand{\ipfscid}[1]{\href{\ipfsgateway#1}{\texttt{\seqsplit{#1}}}}

\newcommand{\magnetlink}[1]{\href{magnet:?xt=urn:btih:#1}{\texttt{\seqsplit{magnet:?xt=urn:btih:#1}}}}

\newcommand{\dockerimage}[2]{%
  \href{#1}{\texttt{#2}}%
}

\ifnonanonymous
    \newcommand{\repoBase}{https://github.com/Erotemic/scatspotter/blob/31f497e2586f0d1560b9bbd65415f9bd36a07585}
\else
    \newcommand{\repoBase}{https://anonymous.4open.science/r/scatspotter-DD67}
\fi

\newcommand{\repolink}[2]{\href{\repoBase/#1}{\texttt{#2}}}

\newcommand{\linkwithtip}[3]{%
  \href{#1}{\pdftooltip{#3}{#2}}%
}

\newcommand{\ghlink}[1]{%
  \href{https://github.com/#1}{#1}
}

\newcommand{\hflink}[1]{%
  \href{https://huggingface.co/#1}{#1}
}

\newcommand{\YOLOPretrained}{
  \linkwithtip
  {https://github.com/WongKinYiu/yolov9/releases/download/v0.1/yolov9-c.pt}
  {sha256 is b66df73be150f1025574b4399148815d5c510cf3d8f7fc7db216228e298132c6}
  {\texttt{v9-c.pt}}
}

\newcommand{\DINOPretrained}{
  \linkwithtip
  {https://github.com/IDEA-Research/GroundingDINO/releases/download/v0.1.0-alpha/groundingdino_swint_ogc.pth}
  {sha256 is 3b3ca2563c77c69f651d7bd133e97139c186df06231157a64c507099c52bc799}
  {\texttt{groundingdino\_swint\_ogc.pth}}
}

\newcommand{\MaskRCNNPretrained}{
  \linkwithtip
  {https://dl.fbaipublicfiles.com/detectron2/ImageNetPretrained/MSRA/R-50.pkl}
  {sha256 is 98f5aabca9d6bcc5d61d0517987356d710a8404e7ffe242caf1d8f343357b448}
  {\texttt{detectron2://ImageNetPretrained/MSRA/R-50.pkl}}
}

\title{``ScatSpotter'' --- A Dog Poop Detection Dataset}

\author{Jonathan Crall\\
Kitware\\
\texttt{jon.crall@kitware.com} \\
}

\begin{document}
\maketitle

\begingroup
\renewcommand{\thefootnote}{}%
\footnotetext{\redact{Accepted to the \emph{International Workshop on Smart Waste Monitoring (WasteVision)} at WACV 2026.}}%
\addtocounter{footnote}{-1}%
\endgroup

\begin{abstract}

Small, amorphous waste objects such as biological droppings and microtrash can
be difficult to see, especially in cluttered scenes, yet they matter for
environmental cleanliness, public health, and autonomous cleanup.  We introduce
``ScatSpotter'': a new dataset of phone images annotated with polygons
around dog feces, collected to train and study object
detection and segmentation systems for small potentially camouflaged outdoor
waste.  We gathered data in mostly urban environments, using a
``before/after/negative'' (BAN) protocol: for a given location, we capture an
image with the object present, an image from the same viewpoint after removal,
and a nearby negative scene that often contains visually similar confusers.

Image collection began in late 2020.  This paper focuses on two dataset
checkpoints from 2025 and 2024.  The dataset contains over 9000
full-resolution images and 6000 polygon annotations.  Of the author-captured
images we held out 691 for validation and used the rest to train.  Via
community participation we obtained a 121-image test set that, while small, is
independent from author-collected images and provides some generalization
confidence across photographers, devices, and locations.  Due to its limited
size, we report both validation and test results.

We explore the difficulty of the dataset  using off-the-shelf VIT, MaskRCNN,
YOLO-v9, and DINO-v2 models.  Zero-shot DINO performs poorly, indicating
limited foundational-model coverage of this category.  Tuned DINO is the best
model with a box-level average precision of 0.69 on a 691-image validation set
and 0.70 on the test set.  These results establish strong baselines and
quantify the remaining difficulty of detecting small, camouflaged waste
objects.

To support open access to models and data (CC-BY 4.0 license), we
compare centralized and decentralized distribution mechanisms and discuss 
trade-offs for sharing scientific data.  Code for
experiments and project details are \href{\repoBase}{hosted on GitHub}.


\end{abstract}


\begin{figure}[t]
\centering
\begin{subfigure}[t]{0.49\textwidth}
    \centering
    \includegraphics[width=\linewidth]{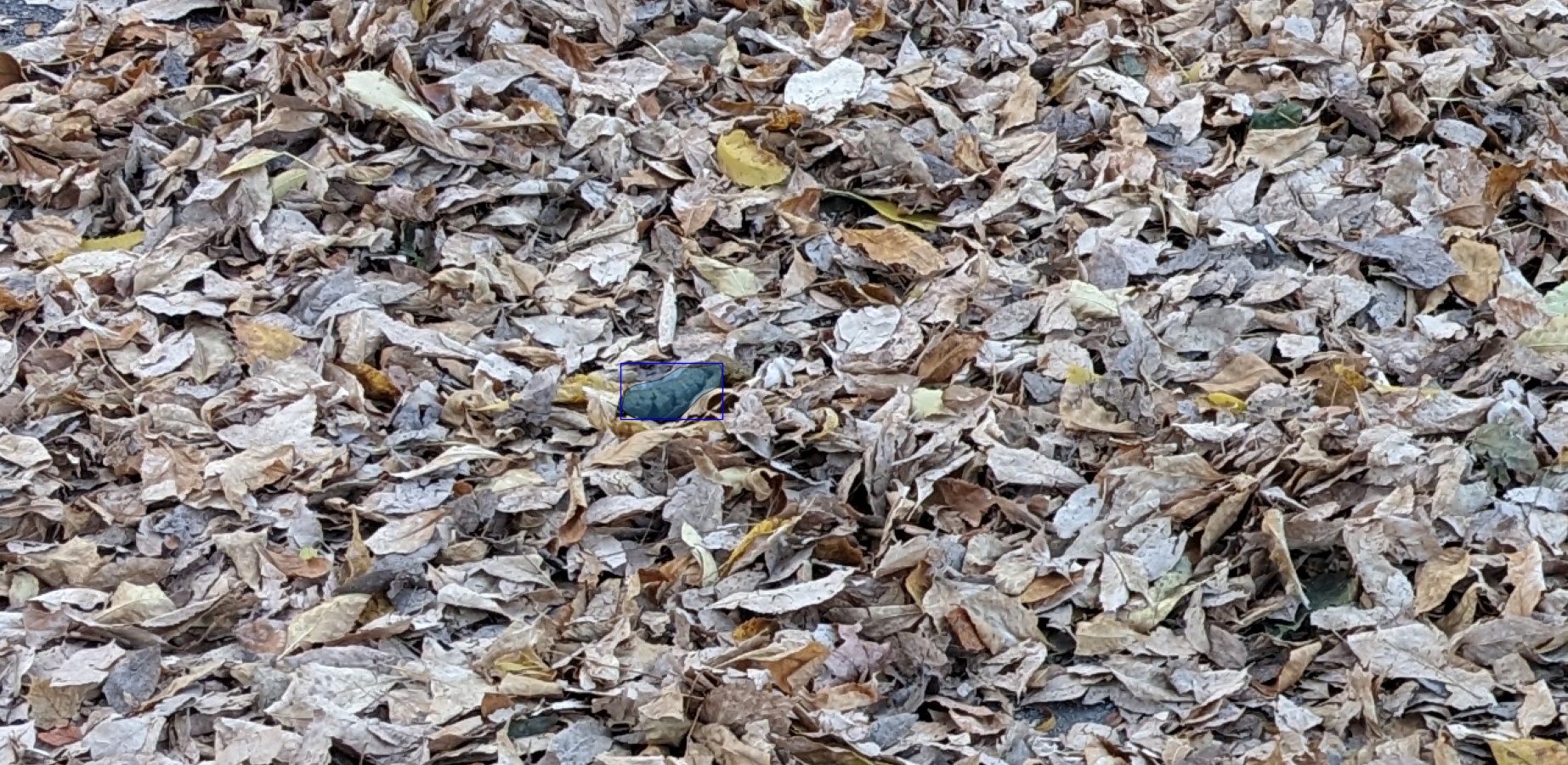}
    \caption[]{
        A zoomed in example of an annotated object in a challenging
        condition: a scene cluttered with leaves. The similarity between the leaves
        and the poop causes a camouflage effect that can make detecting it difficult.
        The poop is highlighted in blue, but in the original image is difficult
        to distinguish.
    }
    \label{fig:HardCase}
\end{subfigure}
\hfill
\begin{subfigure}[t]{0.49\textwidth}
    \centering
    \includegraphics[width=\linewidth]{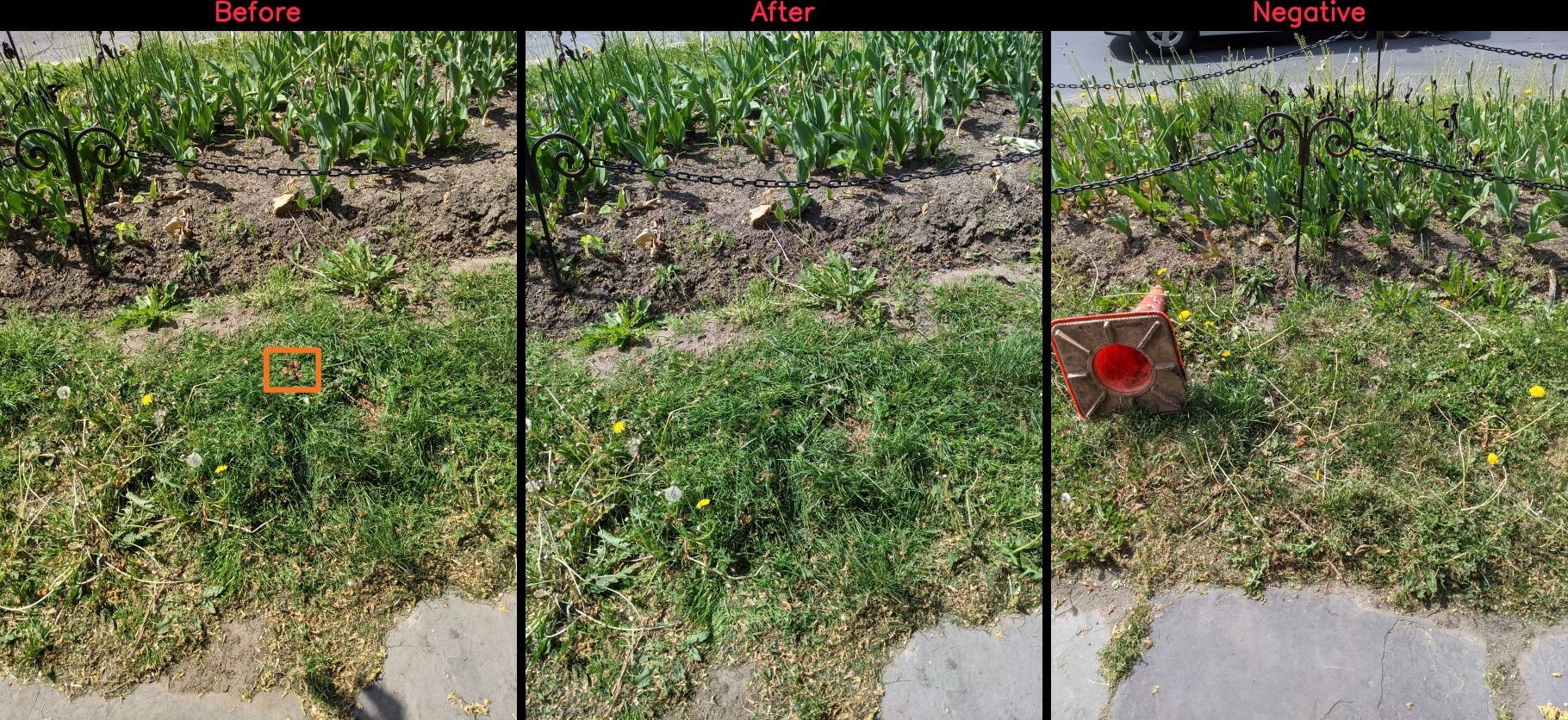}
    \caption[]{
        The ``before/after/negative'' protocol.
        The orange box highlights the location of the poop 
        in the ``before'' image.
        In the ``after'' image, it is the same scene and viewpoint but the poop has been removed.
        The ``negative'' image is a nearby similar scene, potentially with a distractor.
        Note that the object is small relative to the image size.
    }
    \label{fig:ThreeImages}
\end{subfigure}
\caption{(a) A challenging annotation case due to clutter and camouflage. (b) An image triplet from the BAN protocol.}
\label{fig:Combined}
\end{figure}

\section{Introduction}
\label{sec:intro}

\begin{figure*}[t]
\centering
\includegraphics[width=1.0\textwidth]{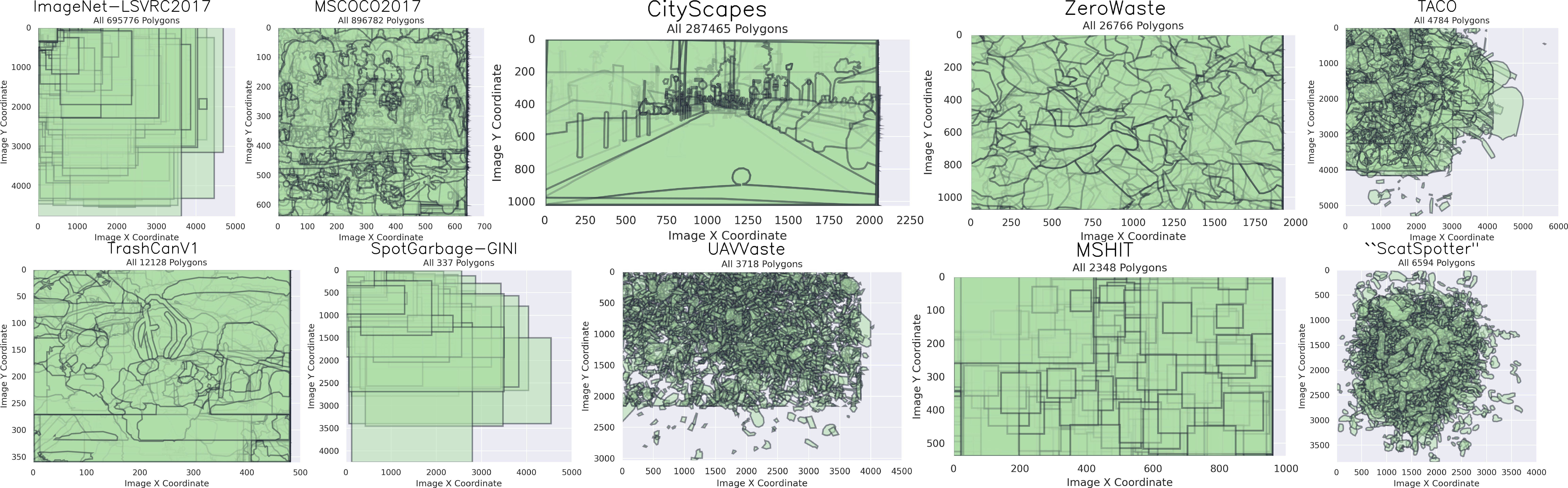}
\caption[]{
    A comparison of all of the annotations for different datasets including ours.
    All polygon annotations drawn in a single plot with $0.8$ opacity to
    demonstrate the distribution in annotation location, shape, and size with
    respect to image coordinates.
}
\label{fig:compare_allannots}
\end{figure*}

\newcommand{\RelatedDatasetCaption}{
    \caption{Related datasets.
    Columns list dataset name, number of categories, images, and annotations.
    Image W $\times{}$ H gives median image dimensions;
    Ann Area$^{0.5}$ is the median square root of annotation area (pixels);
    Size is disk requirements in GB; 
    Annot Type is the labeling method.
    \Cref{fig:compare_allannots} shows the distribution of annotation shapes, sizes, and locations.
    }
    \label{tab:related_datasets}
}
\begin{table*}[t]
\centering
\ifwacv \else \RelatedDatasetCaption \fi
\begin{tabular}{lrrrcrrl}
\toprule
Name & \#Cats & \#Images & \#Annots & \makecell{Image\\W $\times{}$ H} & \makecell{Annot\\Area$^{0.5}$} & \makecell{Disk\\Size} & \makecell{Annot\\Type} \\
\midrule
ImageNet\cite{ILSVRC15}    & 1,000 & 594,546 & 695,776 & 500 $\times{}$ 374 & 239 & 166GB & box \\
MSCOCO\cite{lin_microsoft_2014}      & 80 & 123,287 & 896,782 & 428 $\times{}$ 640 & 57 & 50GB & polygon \\
CityScapes\cite{cordts2015cityscapes}  & 40 & 5,000 & 287,465 & 2,048 $\times{}$ 1,024 & 50 & 78GB & polygon \\
ZeroWaste \cite{bashkirova_zerowaste_2022}   & 4 & 4,503 & 26,766 & 1,920 $\times{}$ 1,080 & 200 & 10GB & polygon \\
TrashCanV1\cite{hong2020trashcansemanticallysegmenteddatasetvisual}  & 22 & 7,212 & 12,128 & 480 $\times{}$ 270 & 54 & 0.61GB & polygon \\
UAVVaste\cite{rs13050965}    & 1 & 772 & 3,718 & 3,840 $\times{}$ 2,160 & 55 & 2.9GB & polygon \\
SpotGarbage\cite{mittal2016spotgarbage} & 1 & 2,512 & 337 & 754 $\times{}$ 754 & 355 & 1.5GB & category \\
TACO\cite{proenca_taco_2020}        & 60 & 1,500 & 4,784 & 2,448 $\times{}$ 3,264 & 119 & 17GB & polygon \\
MSHIT\cite{mshit_2020}       & 2 & 769 & 2,348 & 960 $\times{}$ 540 & 99 & 4GB & box \\
Ours        & 1 & 9,296 & 6,594 & 4,032 $\times{}$ 3,024 & 87 & 60GB & polygon \\
\bottomrule
\end{tabular}
\ifwacv \RelatedDatasetCaption \fi
\end{table*}

\begin{figure*}[t]
\centering
\includegraphics[width=1\textwidth]{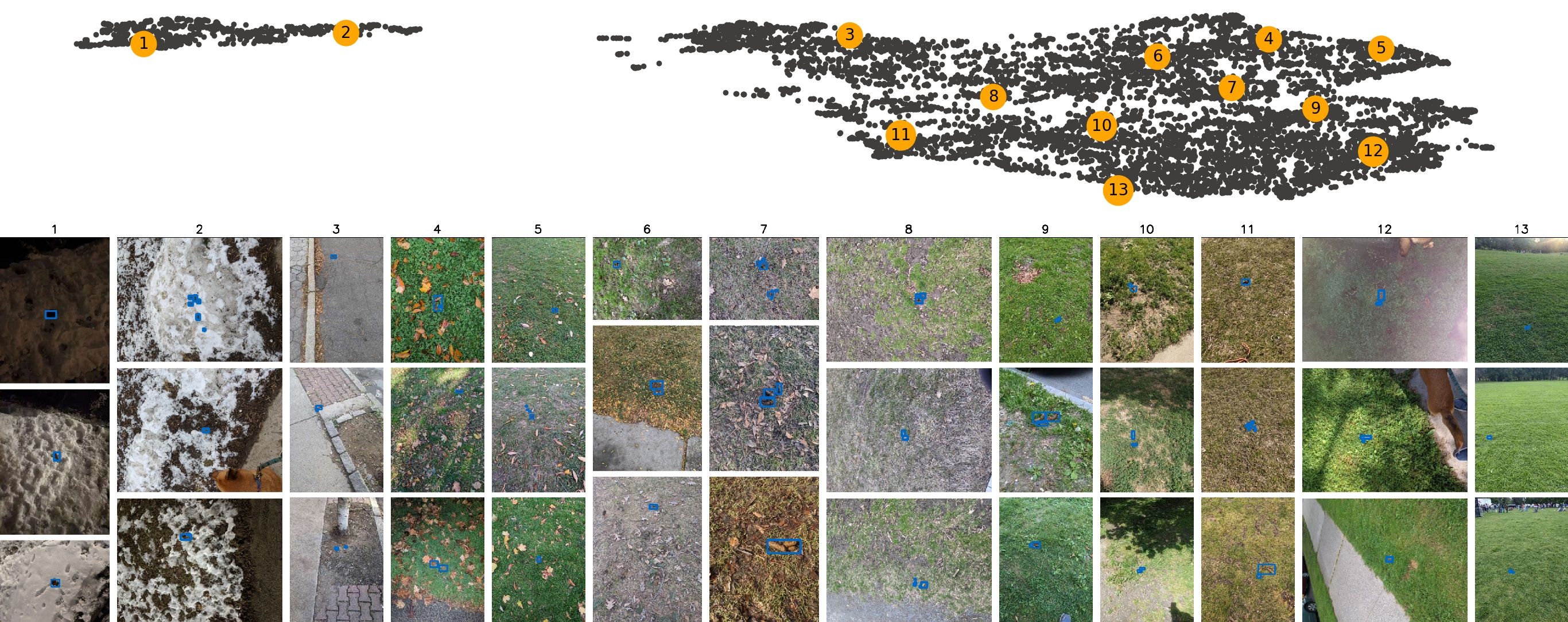}%
\caption[]{
    Example images from 2D UMAP clusters \cite{mcinnes_umap_2020}.
    Each point in the top image represents a 2D-projected embedding, with numbered orange dots indicating nearby
      images in the bottom columns.
    Blue annotation boxes are shown.
    A clear separation emerges between snowy (columns 1-2) and non-snowy images (columns 3-13).
}
\label{fig:umap_dataset_viz}
\end{figure*}

Autonomous and AI-assisted waste monitoring is increasingly achievable with modern object detection and
segmentation methods \cite{sandler_mobilenetv2_2018,
  siam_rtseg_2018,yu_mobilenet_yolo_2023,liu_grounding_2024} combined with large annotated datasets.
Substantial progress has been made in detecting large or conspicuous objects, especially those well represented in foundational training corpora. 
However, small and irregular waste objects --- such as biological droppings or microtrash --- 
are underrepresented in existing datasets and remain difficult to detect.
These objects --- such as the example illustrated in  \Cref{fig:HardCase} ---
are often low contrast, variable in appearance, and confusable with
natural clutter, making them challenging for both humans and vision systems.

To address this gap we introduce a new dataset which, in formal settings, we call ``ScatSpotter''.
Our dataset contains high-resolution images of dog poop in most of which are
from urban, outdoor environments in a single city.
The dataset exhibits variation in appearance, season, lighting, and background clutter despite
  biases toward the author's dogs and geographic region.
Poops are annotated with polygons, making the dataset suitable for both detection and segmentation models.
To assist with annotation and provide counterfactual examples we collect images using a
  ``before/after/negative'' (BAN) protocol as shown in \Cref{fig:ThreeImages}.

One motivating use case, which originally inspired this work, is a phone application that assists dog owners
  in locating their dog's poop in a leafy park for easier cleanup.
Other applications include automated waste disposal to keep sidewalks, parks, and backyards clean, tools for monitoring
  wildlife populations via droppings, and warning systems in smart glasses to prevent people from stepping in
  poop.
Although we focus on a single class, dog poop provides an accessible prototypical example for the
  broader problem of detecting small, amorphous, and often camouflaged waste in outdoor environments --- a
  challenge in common with tasks such as litter detection, microtrash identification, and wildlife monitoring.
The visual difficulty of the domain, rather than the specific species, is the focus of this work.

Beyond the dataset itself, we are also interested in how large datasets can be
shared efficiently and robustly.  Centralized methods such as Girder
\cite{girder_2024} and HuggingFace Datasets \cite{huggingface_datasets} are a
typical choice, offering high speeds, but they can be costly for individuals,
often requiring institutional support or paid hosting services.  They are also
prone to outages and lack built-in data validation.  In contrast, decentralized
methods allow volunteers to host data and offers built-in validation of data
integrity.  This motivates us to compare and contrast BitTorrent
\cite{cohen_incentives_2003}, and IPFS \cite{benet_ipfs_2014} as mechanisms for
distributing datasets.

Our contributions are:
1) A challenging new \textbf{open dataset} of images with polygon annotations for small, camouflaged waste
   objects (using dog poop as a case study).
2) A set of trained \textbf{baseline models}.
3) A \textbf{comparison of dataset distribution} methods.
Together, these contributions are intended to support future work on small-object waste detection, smart
waste monitoring, and environmentally focused computer vision applications.
For F.A.Q., see \Cref{sec:faq}. 


\section{Related Work}
\label{sec:relatedwork}

To the best of our knowledge, our dataset is currently the largest publicly available collection of
  annotated dog poop images, but it is not the first.
A dataset of 100 dog poop images was collected and used to train a FasterRCNN model
  \cite{neeraj_madan_dog_2019} but this dataset and model are not publicly available.
The company iRobot has a dataset of annotated indoor poop images used to train Roomba j7+ to avoid
  collisions \cite{roomba_2021}, but as far as we are aware, this is not available.
In terms of available poop detection datasets we are only aware of MSHIT~\cite{mshit_2020} which is much
  smaller, only contains box annotations, and the objects of interest are plastic toy poops.

Compared to benchmark object localization and segmentation datasets~\cite{ILSVRC15,
  lin_microsoft_2014,cordts2015cityscapes} ours is much smaller and focused only on a single category.
However, when compared to litter and trash datasets
  \cite{bashkirova_zerowaste_2022,proenca_taco_2020,hong2020trashcansemanticallysegmenteddatasetvisual,mittal2016spotgarbage,rs13050965}
  ours is among the largest in terms of number of images / annotations, image size, and total dataset size.
ZeroWaste~\cite{bashkirova_zerowaste_2022} uses a ``before/after'' protocol similar to our BAN protocol.
We provide an overview of these related datasets in \Cref{tab:related_datasets}.
Among all of these, ours stands out for having the highest resolution images and the smallest objects
  relative to that resolution.
For a review of additional waste related datasets, refer to \cite{agnieszka_waste}.

\Cref{sec:dataset_transfer} discusses the logistics and tradeoffs between dataset distribution mechanisms
  with a focus on comparing centralized and decentralized methods.
IPFS~\cite{benet_ipfs_2014} and BitTorrent~\cite{cohen_incentives_2003} are the decentralized 
  mechanisms we evaluate, but others exist such as Secure Scuttlebut \cite{tarr_secure_2019} and Hypercore
  \cite{frazee_dep-0002_nodate}, which we did not test.


\section{Dataset}
\label{sec:dataset}

Our first contribution is the creation of a new open dataset which consists of images of dog poop in mostly
  urban, mostly outdoor environments, from mostly a single city.
The data is annotated to support object detection and segmentation tasks.
The majority of the images feature fresh poop from three specific medium sized dogs, but there are
  a significant number of images with poops of unknown age and from unknown dogs.

Despite these biases, the dataset has significant image variations.
To provide a gist, we computed UMAP \cite{mcinnes_umap_2020} embeddings using ResNet50
  \cite{he2016deep} descriptors, and display images corresponding with clusters in
  \Cref{fig:umap_dataset_viz}.

More details about the dataset are available in a standardized datasheet
\cite{gebru_datasheets_2021} that covers the motivation, composition,
collection, preprocessing, uses, distribution, and maintenance. This is
distributed with the data itself.

\subsection{Dataset Collection}

A single researcher on dog walks photographed fresh dog poop, mostly their own
dogs, but often others. Distance was varied for diversity. Most
images were taken following the ``before/after/negative'' (BAN) protocol.  
A BAN triple comprises a ``before'' shot of the poop, an ``after'' shot
post removal, and a ``negative'' shot of a nearby lookalike (e.g., pine cones,
leaves).  We only use them for negative sampling, but they could enable
contrastive triplet losses \cite{schroff_facenet_2015}.

The majority of images follow the BAN protocol, but there are exceptions.
The first six months of data collection only involved the ``before/after'' part of the protocol. 
We began collecting the third negative image after a colleague suggested it.
In some cases, the researcher failed or was unable to take the second or third image.
These exceptions are often programmatically identifiable.
  
We also received 121 contributor images, mostly outside the BAN protocol, which we use as a test set.
Due to the small size, our main results also include validation scores.

\subsection{Dataset Annotation}

Images were annotated using labelme \cite{wada_labelmeailabelme_nodate}.
Most annotations were initialized using SAM and a point prompt.
All AI polygons were manually reviewed.
In most cases only small manual adjustments were needed, but there were a significant number of cases where
  SAM did not work well and fully manual annotations were needed.
Regions with shadows seemed to cause SAM the most trouble, but there were other failure cases.
Unfortunately, there is no metadata to indicate which polygons were manually created or done using AI.
However, the number of vertices may be a reasonable proxy to estimate this, as polygons generated by SAM
  tend to have higher fidelity boundaries.
The boundaries of the annotated polygons are illustrated in \Cref{fig:compare_allannots}.

Data collected after 2024-07-03 was annotated with the help of models trained
on prior data. Again, all predictions were manually verified or corrected. In
these later cases, false positive annotations were labeled (e.g. stick, leaf),
but because these categories are not labeled exhaustively, we exclude them from
all analysis in this paper.

\begin{figure}[t]
\centering
\begin{subfigure}[t]{0.48\textwidth}
    \centering
    \includegraphics[width=\textwidth]{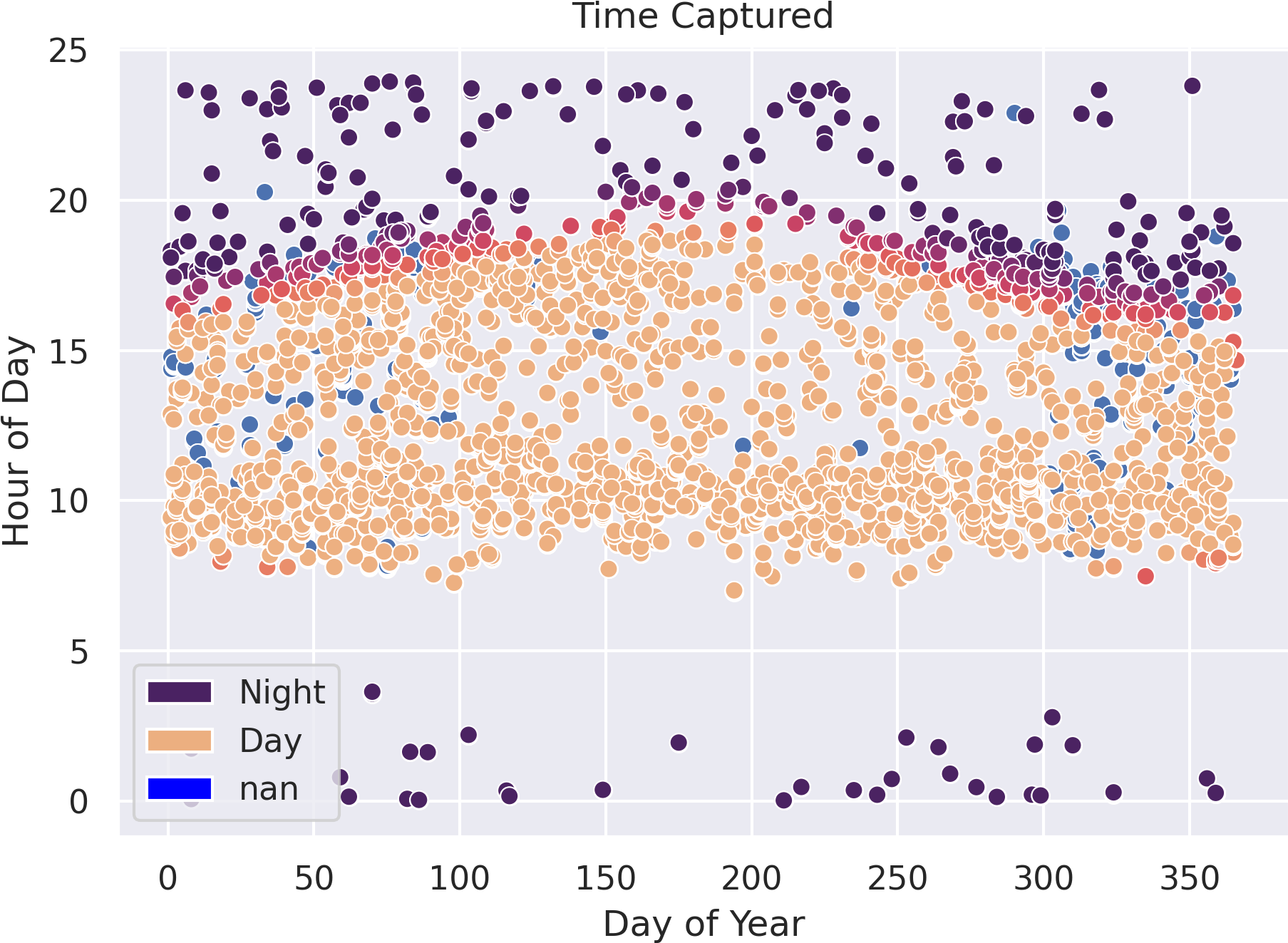}
    \caption{
        The time-of-year vs time-of-day of each image show lighting and seasonal
        variation.  On the x-axis, 0 is January 1st. On the y-axis, 0 is
        midnight.  Color estimates daylight based on location (nan means not available).
        Most images are in the day, but many are at night with flash or long exposure.
    }
    \label{fig:TimeOfDayDistribution}
\end{subfigure}
\hfill
\begin{subfigure}[t]{0.48\textwidth}
    \centering
    \includegraphics[width=\textwidth]{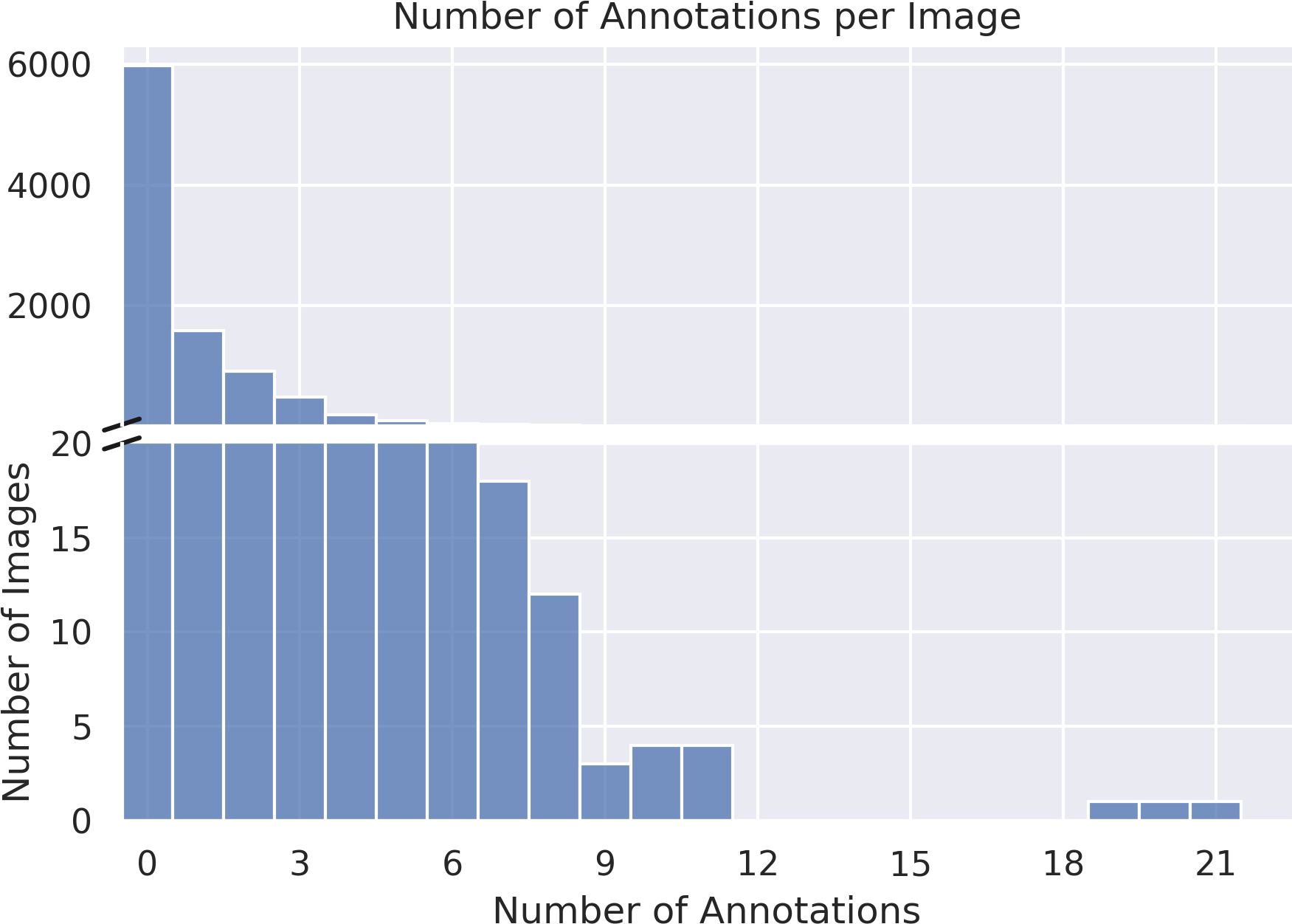}
    \caption{
        The histogram show variation in annotations per image.
        Only 35\% (3,314) of images contain annotations; 65\% (5,982) are known negatives.
        About half of the negatives were taken immediately after pickup; the
        rest are from nearby locations with potential lookalikes. Note the split y-axis.
    }
    \label{fig:AnnotsPerImage}
\end{subfigure}
\caption{Dataset distributions. (a) Time and daylight scatterplot. (b) Annotation count histogram.}
\label{fig:TimeAndAnnots}
\end{figure}

\subsection{Dataset Properties and Statistics}
\label{subsec:datastat}



The data was captured at a regular rate over 4.3 years, primarily in parks and sidewalks within a small
  city.
Weather conditions varied across snowy, sunny, rainy, and foggy.
A visual representation of the distribution of seasons, time-of-day, daylight, and capture rate is provided
  in \Cref{fig:TimeOfDayDistribution}.

The dataset is available in full resolution.
Almost all images were taken using the same phone-camera, with a consistent size of 4,032
  $\times$ 3,024 (up to EXIF rotation).
The images are stored as 8-bit JPEGs with RGB channels, and most include overviews (i.e., image pyramids),
  allowing for fast loading of downscaled versions.

Due to the BAN protocol, about one-third of the images contain
annotations, the rest were taken after the object(s) were removed.  Consequently, most
images have no annotations. When present, annotations are typically singular, but
multiple are common due to:
1) fragmented droppings,
2) dogs pooping together,
3) repeated poops in the same area over time (sometimes hard to distinguish from dirt).
The number of annotations per image is given in \Cref{fig:AnnotsPerImage}.

\subsection{Dataset Splits}

Our dataset is split into training, validation, and test sets based on the year and day of image capture and
  photographer.
Only data captured by the authors is used for training and validation.
Of these, images from 2021-2023, 2025 and beyond are assigned to the training set. 
Images from 2020 are used for
  validation.
For data from 2024, we consider the ordinal date $n$ of each image and include it in the validation set if
  $n \equiv 0 \ (\textrm{mod}\ 3)$; otherwise, it is assigned to training.

For testing data, we use contributor images to not bias our results based on the way the authors took
  images.
These splits are provided in the COCO JSON format \cite{lin_microsoft_2014} as well as a WebDataset
  \cite{huggingfacewebdataset} on HuggingFace.

\section{Baseline Models}
\label{sec:models}

As our second contribution, we trained and evaluated models to establish a baseline for future comparisons.
Specifically we train 7 model variants.
We trained a semantic segmentation vision transformer variant (VIT-sseg-s)
  \cite{Greenwell_2024_WACV,crall_geowatch_2024}, which was only trained from scratch.
We trained two MaskRCNN \cite{he2017mask} models (specifically the \texttt{R\_50\_FPN\_3x} configuration),
  one starting from pretrained ImageNet weights (MaskRCNN-p), and one starting from scratch
  (MaskRCNN-s).
Similarly we trained YOLO-v9 \cite{wang2024yolov9} both from scratch and using pretrained ImageNet weights.
Lastly, we evaluated the foundational Grounding DINO \cite{liu_grounding_2024} model. 
In the zero-shot setting we used \texttt{IDEA-Research/grounding-dino-tiny} using the prompt: "animalfeces".
Finally, we fine-tune evaluate the same DINO model using \cite{OpenGroundingDino}.

The number of parameters for MaskRCNN, VIT, GroundingDINO, and YOLO are 44M, 26M, 172M, and 51M.
Hyperparameters are given in \Cref{sec:experiment_details}.

For these baseline models, the training data was limited to an older subset taken before 2024-07-03.
Our training dataset consists of 5,747 images and is identified by a suffix of {\tt 1e73d54f}, which is the
  prefix of its content hash.
The validation set contains 691 images and has a suffix of {\tt 99b22ad0}.
The test set, consists of the 121 images, has a suffix of {\tt 6cb3b6ff}, and includes contributor images
  up to 2025-04-20.
The evaluated models were selected based on their Box-AP validation scores.

The primary detection ``Box'' evaluation computes standard COCO object detection metrics
  \cite{lin_microsoft_2014}.
MaskRCNN, GroundingDINO, and YOLO-v9 natively output scored bounding boxes, but for the VIT-sseg model, we convert heatmaps into boxes
  by thresholding the probability maps and taking the extend of the resulting polygons as bounding
  boxes.
The score is taken as the average heatmap response under the polygon.
Bounding box evaluation has the advantage that small and large annotations contribute equally to the score,
  but it can also be misleading for datasets where the notion of an object instance can be ambiguous.

To complement the box evaluation, we performed a pixelwise evaluation, which is more sensitive to the
  details of the segmented masks, but also can be biased towards larger annotations with more pixels.
The corresponding truth and predicted pixels were accumulated into a confusion matrix, allowing us to
  compute standard metrics \cite{powers_evaluation_2011} such as precision, recall, false positive rate, etc.
For the VIT-sseg model, computing this score is straightforward, but for MaskRCNN we accumulate per-box
  heatmaps into a larger full image heatmap, which can then be scored.
Because YOLO-v9 and GroundingDINO do not produce masks, they were excluded from pixelwise evaluation.

Quantitative results for each of these models on box and pixel metrics are shown in
  \Cref{tab:model_results}.
Because the independent test set is only 121 images, we also present results on the larger validation
  dataset.
Corresponding validation results are illustrated in \Cref{fig:vali_results_all_models} and
test results in \Cref{fig:test_results_all_models}.


\newcommand{\tb}[1]{\textbf{#1}}

\newcommand{\ResultCaption}{
\caption[]{
    Baseline model performance on validation and test sets.
    Suffixes indicate training conditions: -p (pretrained), -s (scratch), -t (tuned), -z (zero-shot).
    Metrics include box- and pixel-level AP (area under precision-recall), AUC (area under ROC), F1, and TPR (recall),
    computed using scikit-learn \cite{scikit-learn}.
    Pretrained models outperform with the tuned foundational Grounding DINO model performing best.
    Note: VIT-sseg was tuned more extensively and operated at full resolution,
    while MaskRCNN, DINO, and YOLO used off-the-shelf settings (that resized
    images) and may improve with additional tuning.
}
\label{tab:model_results}
}

\begin{table*}[t]
\ifwacv \else \ResultCaption \fi
\centering

    \begin{subtable}[b]{\textwidth} 
    \caption{Validation (n=691)}
    \centering
        \begin{tabular}{lllllllll}
        \toprule
         Model           & \makecell{AP\\Box}   & \makecell{AUC\\Box}   & \makecell{F1\\Box}   & \makecell{TPR\\Box}   & \makecell{AP\\Pixel}   & \makecell{AUC\\Pixel}   & \makecell{F1\\Pixel}   & \makecell{TPR\\Pixel}   \\
        \midrule
         MaskRCNN-p      & 0.61                 & \textbf{0.72}         & 0.62                 & 0.57                  & 0.74                   & 0.91                    & \textbf{0.74}          & 0.68                    \\
         MaskRCNN-s      & 0.26                 & 0.58                  & 0.35                 & 0.31                  & 0.43                   & 0.89                    & 0.48                   & 0.50                    \\
         VIT-sseg-s      & 0.48                 & 0.53                  & 0.60                 & 0.51                  & \textbf{0.76}          & \textbf{0.97}           & 0.74                   & \textbf{0.69}           \\
         GroundingDINO-t & \textbf{0.69}        & 0.63                  & \textbf{0.74}        & \textbf{0.68}         & --                     & --                      & --                     & --                      \\
         GroundingDINO-z & 0.08                 & 0.21                  & 0.20                 & 0.25                  & --                     & --                      & --                     & --                      \\
         YOLO-v9-p       & 0.41                 & 0.59                  & 0.50                 & 0.42                  & --                     & --                      & --                     & --                      \\
         YOLO-v9-s       & 0.33                 & 0.41                  & 0.44                 & 0.37                  & --                     & --                      & --                     & --                      \\
        \bottomrule
        \end{tabular}
    \end{subtable}

  \hfill 

    \begin{subtable}[b]{\textwidth} 
    \caption{Test (n=121)}
    \centering
        \begin{tabular}{lllllllll}
        \toprule
         Model      & \makecell{AP\\Box}   & \makecell{AUC\\Box}   & \makecell{F1\\Box}   & \makecell{TPR\\Box}   & \makecell{AP\\Pixel}   & \makecell{AUC\\Pixel}   & \makecell{F1\\Pixel}   & \makecell{TPR\\Pixel}     \\
        \midrule
         MaskRCNN-p      & 0.61          & \textbf{0.70} & 0.65          & 0.60          & \textbf{0.81} & \textbf{0.85} & \textbf{0.78} & \textbf{0.73} \\
         MaskRCNN-s      & 0.25          & 0.47          & 0.34          & 0.30          & 0.39          & 0.80          & 0.41          & 0.44          \\
         VIT-sseg-s      & 0.39          & 0.40          & 0.52          & 0.41          & 0.41          & 0.82          & 0.48          & 0.37          \\
         GroundingDINO-t & \textbf{0.70} & 0.67          & \textbf{0.76} & \textbf{0.68} & --            & --            & --            & --            \\
         GroundingDINO-z & 0.23          & 0.30          & 0.39          & 0.38          & --            & --            & --            & --            \\
         YOLO-v9-p       & 0.44          & 0.55          & 0.51          & 0.50          & --            & --            & --            & --            \\
         YOLO-v9-s       & 0.36          & 0.36          & 0.48          & 0.37          & --            & --            & --            & --            \\
        \bottomrule
        \end{tabular}
    \end{subtable}

\ifwacv \ResultCaption \fi
\end{table*}

\definecolor{fpred}{HTML}{F42836} 
\definecolor{tppred}{HTML}{0068C7} 
\definecolor{fntrue}{HTML}{800080} 
\definecolor{tptrue}{HTML}{3EAE2B} 
\definecolor{neutral}{HTML}{242A37} 

\newcommand{\FP}{\textcolor{fpred}{false positive}}
\newcommand{\TPpred}{\textcolor{tppred}{true-positive prediction}}
\newcommand{\FN}{\textcolor{fntrue}{false negative}}
\newcommand{\TPtrue}{\textcolor{tptrue}{true positive (GT)}}
\newcommand{\TN}{\textcolor{neutral}{true negative}}

\begin{figure*}[ht]
\centering
\includegraphics[width=1.0\textwidth]{figures/agg_viz_results2/vali_imgs691_99b22ad0.kwcoco/results_geowatch-scratch_heatmap_confusion_components.jpg}%
\hfill
(a) VIT-sseg-scratch (validation set results)
\includegraphics[width=1.0\textwidth]{figures/agg_viz_results2/vali_imgs691_99b22ad0.kwcoco/results_detectron-pretrained_heatmap_confusion_components.jpg}%
\hfill
(b) MaskRCNN-pretrained (validation set results)
\includegraphics[width=1.0\textwidth]{figures/agg_viz_results2/vali_imgs691_99b22ad0.kwcoco/results_detectron-scratch_heatmap_confusion_components.jpg}%
\hfill
(c) MaskRCNN-scratch (validation set results)
\includegraphics[width=1.0\textwidth]{figures/agg_viz_results2/vali_imgs691_99b22ad0.kwcoco/results_yolo_v9-scratch_detection_confusion.jpg}%
\hfill
(d) YOLO-v9-scratch (validation set results)
\includegraphics[width=1.0\textwidth]{figures/agg_viz_results2/vali_imgs691_99b22ad0.kwcoco/results_yolo_v9-pretrained_detection_confusion.jpg}%
\hfill
(e) YOLO-v9-pretrained (validation set results)
\includegraphics[width=1.0\textwidth]{figures/agg_viz_results2/vali_imgs691_99b22ad0.kwcoco/results_grounding_dino-zero_detection_confusion.jpg}%
\hfill
(f) GroundingDINO-zero-shot (validation set results)
\includegraphics[width=1.0\textwidth]{figures/agg_viz_results2/vali_imgs691_99b22ad0.kwcoco/results_grounding_dino-tuned_detection_confusion.jpg}%
\hfill
(g) GroundingDINO-tuned (validation set results)
\includegraphics[width=1.0\textwidth]{figures/agg_viz_results2/vali_imgs691_99b22ad0.kwcoco/results_input_images.jpg}%
\hfill
(h) Inputs from the validation set
\caption[]{
    Qualitative results from validation-selected models applied to the same validation images.
    Subfigures (a-c) show results for VIT and MaskRCNN, including both the binarized classification map 
    (\textcolor{tptrue}{true positives in green}, 
     \textcolor{fpred}{false positives in red}, 
     \textcolor{fntrue}{false negatives in purple}, 
     \textcolor{neutral}{true negatives in black}) 
    and the predicted heatmap before binarization.  
    Subfigures (d-g) show bounding-box detections from YOLO-v9 and Grounding DINO, using the same color scheme 
    (\textcolor{tppred}{blue = true-positive predicted boxes}; 
     \textcolor{tptrue}{green = matched ground truth}).
    Subfigure (h) shows the input image.
}
\label{fig:vali_results_all_models}
\end{figure*}

\begin{figure*}[ht]
\centering
\includegraphics[width=1.0\textwidth]{figures/agg_viz_results2/test_imgs121_6cb3b6ff.kwcoco/results_geowatch-scratch_heatmap_confusion_components.jpg}%
\hfill
(a) VIT-sseg-scratch (test set results)
\includegraphics[width=1.0\textwidth]{figures/agg_viz_results2/test_imgs121_6cb3b6ff.kwcoco/results_detectron-pretrained_heatmap_confusion_components.jpg}%
\hfill
(b) MaskRCNN-pretrained (test set results)
\includegraphics[width=1.0\textwidth]{figures/agg_viz_results2/test_imgs121_6cb3b6ff.kwcoco/results_detectron-scratch_heatmap_confusion_components.jpg}%
\hfill
(c) MaskRCNN-scratch (test set results)
\includegraphics[width=1.0\textwidth]{figures/agg_viz_results2/test_imgs121_6cb3b6ff.kwcoco/results_yolo_v9-scratch_detection_confusion.jpg}%
\hfill
(d) YOLO-v9-scratch (test set results)
\includegraphics[width=1.0\textwidth]{figures/agg_viz_results2/test_imgs121_6cb3b6ff.kwcoco/results_yolo_v9-pretrained_detection_confusion.jpg}%
\hfill
(e) YOLO-v9-pretrained (test set results)
\includegraphics[width=1.0\textwidth]{figures/agg_viz_results2/test_imgs121_6cb3b6ff.kwcoco/results_grounding_dino-zero_detection_confusion.jpg}%
\hfill
(f) GroundingDINO-zero-shot (test set results)
\includegraphics[width=1.0\textwidth]{figures/agg_viz_results2/test_imgs121_6cb3b6ff.kwcoco/results_grounding_dino-tuned_detection_confusion.jpg}%
\hfill
(g) GroundingDINO-tuned (test set results)
\includegraphics[width=1.0\textwidth]{figures/agg_viz_results2/test_imgs121_6cb3b6ff.kwcoco/results_input_images.jpg}%
\hfill
(h) Inputs from the test set
\caption[]{
    Qualitative results from validation-selected models applied to test images.
    Subfigures (a-c) show results for VIT and MaskRCNN, including both the binarized classification map
    (\textcolor{tptrue}{true positives in green}, 
     \textcolor{fpred}{false positives in red}, 
     \textcolor{fntrue}{false negatives in purple}, 
     \textcolor{neutral}{true negatives in black}) 
    and the predicted heatmap before binarization.  
    Subfigures (d-g) show bounding-box detections from YOLO-v9 and Grounding DINO, using the same color scheme 
    (\textcolor{tppred}{blue = true-positive predicted boxes}; 
     \textcolor{tptrue}{green = matched ground truth}).
    Subfigure (h) shows the input image.
}
\label{fig:test_results_all_models}
\end{figure*}

All models were trained on a single machine with an Intel Core i9-11900K CPU and an NVIDIA GeForce RTX 3090 GPU. 
Our environmental impact
\footnote{Over all of our experimentation, prediction and evaluation took 14 days, consuming 108 kWh and emitting 23~\cotwo~kg (CodeCarbon \cite{lacoste2019codecarbon}). Training was estimated at 164 days and 1359 kWh, yielding 285~\cotwo~kg, assuming a 345W GPU draw and a 0.21~$\frac{\textrm{kg}\cotwo{}}{\textrm{kWh}}$ emission factor. At $\$0.16$/kWh and $\$25$/tonne~\cotwo, total cost was \$242.37. More details in \Cref{sec:general_environmental_impact}.}
was manageable.


\section{Dataset Transfer Experiment}
\label{sec:dataset_transfer}


Our third contribution is an experiment that studies transfer rates of decentralized and centralized data
  distribution methods. 
For centralized distribution, we use a self-hosted instance of Girder~\cite{girder_2024} and the HuggingFace
  datasets~\cite{huggingface_datasets} platform.
For decentralized clients, we use Transmission~\cite{transmission_2024} (BitTorrent) and
  Kubo~\cite{ipfskubo_2024} (IPFS).
As a baseline, we also measure direct transfers using Rsync~\cite{rsyncprojectrsync_2024}.

For data transfer experiments, we use the 2024-07-03 version of the dataset.
This is content-addressed with the IPFS CID (content identifier):
\ipfscid{bafybeiedwp2zvmdyb2c2axrcl455xfbv2mgdbhgkc3dile4dftiimwth2y}.
The torrent magnet URL is:
\magnetlink{ee8d2c87a39ea9bfe48bef7eb4ca12eb68852c49},
and is tracked on Academic Torrents \cite{academic_torrents_Cohen2014}.
More details in \Cref{sec:datset_discuss}.




The HuggingFace results stand out, as they are faster than rsync.
We believe this is due to an optimized client and content delivery networks, utilizing CAKE
  \cite{hoiland2018piece} to minimize buffer bloat \cite{gettys2012bufferbloat}.
However, this speed relies on costly centralized infrastructure.
The expected speed from a more modest centralized service is $\sim\!20\times$ slower.

There is an additional $\sim\!4\times$  slowdown between compressed and uncompressed rsync baselines, which needs to be
  considered when comparing decentralized results.
The minimum time column shows that decentralized methods can be competitive with rsync, but on
  average decentralized mechanisms are significantly slower and can be stifled by long peer-discovery times.

\begin{table}[hb]
\centering
\setlength{\tabcolsep}{3pt} 
\begin{tabular}{lcrrrr}
\toprule
Method       & Zipped & $\mu$     & $\sigma$ & Min    & Max     \\
\midrule       
BitTorrent   & No         & 8.36h     & 5.16h    & 2.21h  & 14.39h  \\
IPFS         & No         & 10.68h    & 9.54h    & 1.80h  & 24.62h  \\
Rsync        & No         & 4.84h     & 1.39h    & 3.10h  & 6.10h   \\
Girder       & Yes        & 2.85h     & 2.31h    & 1.05h  & 6.24h   \\
HuggingFace  & Yes        & \bf{0.14h}& 0.03h    & 0.11h  & 0.18h   \\
Rsync        & Yes        & 1.10h     & 0.03h    & 1.07h  & 1.13h   \\
\bottomrule
\end{tabular}
\caption{
Transfer times (in hours) for our 42GB dataset: trials (n), mean ($\mu$), std ($\sigma$).
Each experiment was run 5 times.
Uncompressed transfers provide granular access to individual files, while compressed (zipped) transfers are faster.
}
\label{tab:transfertime}
\end{table}


\section{Conclusion}

We have introduced the largest open dataset of high resolution images with polygon segmentations of dog
  poop.
While only focused on a single class, it is prototypical of challenges that arise in small-waste detection
  relevant to waste monitoring, pollution tracking, and environmental surveillance.
The dataset includes amorphous objects, occlusion, multi-season variation, difficult distractors, daytime /
  nighttime variation.
We have described the dataset collection and annotation process and reported statistics on the dataset.

We provided a recommended train/validation/test split of the dataset, and trained baseline segmentation
  models that perform well, but could likely be improved.
In addition to providing quantitative and qualitative results of the models, we also estimate the resources
  required to perform these training, prediction, and evaluation experiments.

We have published our data and models under a permissive license, and made them available through both
  centralized (Girder and HuggingFace) and decentralized (BitTorrent and IPFS) mechanisms.
Decentralized methods are robust, but suffer from significant network transfer overhead.
HuggingFace has exceptionally fast transfer speeds, has some decentralized
  properties, but lacks content identifiers.


Our dataset enables applications such as mobile feces detection, urban cleanliness monitoring, and
  augmented-reality collision warnings.
Because it trains models to recognize small, irregular, low-contrast objects in cluttered scenes, we predict
  that including ``ScatSpotter'' in foundational training corpora will improve robustness to camouflage and
  small-object ambiguity in a broad range of ecological and waste-monitoring downstream tasks.

  

\ifuseacknowledgement
\section{Acknowledgements}
We would like to thank all of the dogs that produced subject matter, all of the contributors for helping to
  construct the test set, and \redact{Anthony Hoogs} for several suggestions including taking the third
  negative picture.
This work is dedicated to \redact{Bezoar}, a weird and good girl; \redact{Honey}, a sweet red-fox lookalike;
  and \redact{Roadie}, a vicious soft-faced cuddlebug.

\fi

\FloatBarrier

{\small
\bibliographystyle{ieeenat_fullname}
\bibliography{citations}
}

\ifuseappendix
\appendix

\section{Frequently Asked Questions (F.A.Q.)}
\label{sec:faq}

\begin{enumerate}

\item \textbf{Why is the independent test set so small, and is it sufficient to evaluate generalization?}\\
The independent test set contains 121 images that were captured by non-authors but annotated by the primary author. Its current size is primarily a consequence of how many suitable contributions we received so far; we would prefer a larger test set and are actively seeking to grow it. In parallel, we are experimenting with additional feces-related image collections hosted on public platforms (e.g., community datasets on sites such as Roboflow). These external datasets were not part of the original design of ScatSpotter, and they raise separate licensing, attribution, and quality-control questions, so they are not included in the core benchmark or main tables in this paper.\\
The current contributor test set is strictly photographer-disjoint from the main author data and is meant to provide an initial, realistic check on \emph{cross-photographer} generalization, while the larger author-collected data drive training and validation. However, due to its limited size, it should be viewed as an informative but limited indicator of generalization across larger geographic distribution shifts (\eg{} the dataset does not contain many examples of sandy environments).

\vspace{0.6em}
\item \textbf{Why focus on a single class (dog feces)? Is this problem non-trivial?}\\
The benchmark is intentionally single-class. We prioritize high-quality annotations for one difficult, under-served class instead of many shallow labels. Dog feces are extremely common in real-world settings (parks, sidewalks, trails) but often \emph{not} visually obvious, especially in cluttered foliage, snow, or low light. In practice, even attentive owners sometimes lose track of where their dog went, and visually locating poop is a genuine nuisance task.\\
Although the topic is niche, the computer-vision problem is not trivial: the class is small, highly imbalanced, and strongly confounded by leaves, sticks, rocks, mud, stains, and other background clutter. Learning to solve this requires fine-grained texture and shape cues that are also relevant for broader bio-waste detection, environmental contamination, and aspects of animal behavior monitoring.

\vspace{0.6em}
\item \textbf{Given the narrow, single-class focus and limited geographic diversity, how far can results on ScatSpotter be generalized?}\\
We do not claim ScatSpotter is globally representative. Training and validation data are dominated by a single photographer, a small set of dogs, and a specific geographic region. The contributor test set partially offsets this by introducing different users and capture styles, but the overall benchmark remains biased toward the author's environment and devices. We view ScatSpotter as a focused, well-annotated base domain that can be extended with additional classes (e.g., other waste types) and combined with other datasets for new cities, species, or devices. As with any dataset, careful evaluation on the target domain remains important. 

\vspace{0.6em}
\item \textbf{How were the splits made (train/validation/test), and how does this affect overfitting and generalization?}\\
Initially, the validation set consisted of the earliest batch of collected images, with the remainder used for training. This introduced temporal and domain shift by design. As the dataset grew, we found the validation set too small, so we augmented it with images sampled periodically (e.g., every third day in a given year). This increased its size and eliminated the chance that images from the same walk appear in both train and validation, which could artificially inflate performance.\\
The independent test set is composed of contributor images only and does not include negative-only or ``after'' frames. This setup encourages models to generalize across photographers and capture styles rather than overfit to the main author's habits and devices.

\vspace{0.6em}
\item \textbf{What exactly is the ``before/after/negative'' (BAN) protocol, and how are these triplets represented? Were multiple viewpoints of the same scene captured?}\\
Many examples are captured as short sequences: a \emph{before} image (poop present), an \emph{after} image (poop removed, with a best-effort similar viewpoint), and sometimes a later \emph{negative} image of the same area. This began as a practical annotation aid: flipping between images helps the annotator spot small or camouflaged instances. It also naturally produces multiple viewpoints and time offsets of the same scene, which are valuable for studying hard negatives, change detection, and contrastive objectives.\\
In our code, candidate BAN groups are identified using temporal proximity and image matching (e.g., SIFT-based alignment) and are tagged accordingly. In current releases, BAN roles are derived by the provided scripts rather than stored as static fields in the annotation file; future versions will expose these roles as explicit metadata so downstream work can more easily leverage them.

In our code, candidate BAN groups are identified using temporal proximity and image matching (e.g., SIFT-based alignment) and are tagged accordingly. In current releases, BAN roles are derived by the provided scripts rather than stored as static fields in the annotation file; future versions will expose these roles as explicit metadata so downstream work can more easily leverage them. More generally, images with no polygons labeled ``poop'' are implicitly negative examples, but the current version does not make them explicit.

\vspace{0.6em}
\item \textbf{How were annotations produced and validated?}\\
    Most polygon annotations were drawn and reviewed by a single annotator (the primary author). In some cases, detector outputs (e.g., from YOLO or MaskRCNN) provided initial proposals that were then edited, but we do not store explicit flags indicating which polygons were model-seeded. All annotations pass through the same human review process. When the annotator cannot confidently determine whether a region is poop (e.g., old stains, heavily decomposed material), it is labeled as ``unknown'' or ``ignore'' rather than forced into positive or background. The independent contributor test set is annotated by non-authors; systematic differences in their style give a realistic view of how models handle non-author imagery. 

\vspace{0.6em}
\item \textbf{What else is in the images?}\\

Although the benchmark focuses on dog feces, the scenes contain a wide variety of natural clutter and small objects, including leaves, sticks, rocks, pine cones, helicopter seeds, tree bark, grass patches, dirt, occasional microtrash (e.g., bottle caps, cigarette butts), and other incidental elements (e.g., shadows, roots, mulch, acorns, mice). Some of these appear as sparse labels (often when they caused false positives or were otherwise interesting), but they are not exhaustively annotated across the dataset. Many poop instances also carry free-form description tags (e.g., \texttt{fresh}, \texttt{old}, \texttt{crumbly}, \texttt{diarrhea}, \texttt{messy}, \texttt{occluded}, \texttt{camoflauged}, \texttt{snowcovered}, \texttt{unknown}/\texttt{unsure}). For some interesting cases, additional text tags were added to annotations with the idea that in the future VLMs could use them as truth anchors and help propagate text labels to other similar cases (\eg{} ``old'', ``fresh'', ``sick''). These auxiliary tags are potentially useful signals, but they are not yet standardized enough to define separate benchmark tasks. These extra labels are excluded from stats reported in the main text.

\vspace{0.6em}
\item \textbf{What is the primary task in this benchmark (detection or segmentation), and how should I interpret the baseline models and metrics?}\\
The primary task is \emph{object detection}: find and localize poop instances in an image. We annotate polygons for each instance and derive bounding boxes from these masks for use with commonly used detectors. The dense masks also support segmentation models and more detailed error analysis, but segmentation is secondary to detection in the current benchmark.\\
The baseline suite is intentionally heterogeneous: it includes commonly used models in practice (e.g., YOLO-style one-stage detectors, MaskRCNN-style two-stage detectors), segmentation networks (e.g., ViT-based), and open-vocabulary/foundation detectors (e.g., GroundingDINO), each in a configuration that is standard and well supported for that architecture. Different models therefore run at different input resolutions and capacities. We do not claim these baselines are fully optimized or perfectly comparable on every axis; they should be treated as reasonable reference points, not a definitive ranking.\\
We report detection metrics such as AP and AUC, and also F1, IoU, recall/TPR, and precision, which are important in this highly imbalanced regime. For thresholded metrics we select a single global threshold per model on the validation set and reuse it on the test set, unless otherwise noted. Pixel-level metrics are only reported for architectures that produce masks; detector-only models (e.g., some YOLO variants, GroundingDINO in its standard form) do not output per-pixel predictions.

\vspace{0.6em}
\item \textbf{Why include relatively few baseline architectures, and why not emphasize more lightweight or mobile models?}\\
In an ideal world, we would run and maintain a large zoo of models at every evolution of the dataset. In practice, the current ecosystem for robust, large-scale benchmarking of object detection does not yet provide easy, off-the-shelf solutions for doing this in a maintainable way. Our effort in this work was focused primarily on building, curating, and documenting the dataset. As a result, we select a small, diverse set of baselines that cover key families (two-stage detectors, one-stage YOLO-style models, dense segmenters, foundation/open-vocabulary detectors). Some of these are reasonably lightweight and relevant for on-device or robotic deployment; others are heavier and are used both as strong baselines and as annotation assistants. We would like to expand the benchmark suite as better, less ad-hoc benchmarking frameworks become available, and we place relatively less weight on the specific baseline roster in this initial dataset paper.

\vspace{0.6em}
\item \textbf{What is the real benefit of content-addressable distribution and hash-verifiable data here?}\\
Traditional dataset hosting often relies on a stable URL whose contents may change silently over time, requiring implicit trust in the host. In contrast, content-addressable storage (e.g., IPFS CIDs, torrent magnet hashes, hashed annotation files) ties each version of the dataset to a cryptographic digest of its exact contents. If the bits change, the identifier changes. This has two main benefits:
\begin{enumerate}
    \item It becomes easy to verify that your local copy matches the one used in a given paper, by checking the published hashes and simple dataset statistics.
    \item It becomes hard to accidentally or silently modify the dataset without producing a new identifier, which helps keep future extensions and bug fixes honest and traceable.
\end{enumerate}
We still provide user-friendly access via platforms such as Hugging Face, but we encourage scripts and papers to reference the content hashes so that experiments remain reproducible even if hosting providers or URLs change.

\item \textbf{What privacy considerations apply to this dataset, including EXIF metadata?}\\
Some images in the dataset include EXIF metadata (e.g., camera parameters and timestamps). A subset of images---primarily those captured by the author---include GPS location data in their original form. In cases with privacy concerns, we selectively remove sensitive EXIF fields prior to public release.\\
When consent is obtained, encrypted artifacts are distributed. These allow for rehydration of the original metadata by trusted parties with access to the decryption key. \\
Because dataset snapshots are distributed via content-addressed and peer-to-peer mechanisms (\eg{} IPFS and BitTorrent), users should treat published snapshots as immutable.

\end{enumerate}

\section{Dataset}

\subsection{Additional Comparisons}
\label{sec:expanded_relatedwork}

In \Cref{sec:relatedwork} we compared to related work. Here we expand on this
by comparing our analysis plots. Every dataset is converted into the COCO
format and visualized using the same logic. \Cref{fig:compare_allannots}
visualizes the annotations of all datasets. We make similar visualizations 
for other comparable dataset metrics.
\Cref{fig:combo_anns_per_image_histogram_splity} shows the number of annotations per image.
\Cref{fig:combo_image_size_scatter} shows of image sizes in each dataset.
\Cref{fig:combo_obox_size_distribution_logscale} shows the distribution of width and heights of oriented bounding boxes fit to annotation polygons.
\Cref{fig:combo_polygon_area_vs_num_verts_jointplot} shows the area of each polygon versus the number of vertices (which could be used to estimate the likelihood a polygon was generated by AI for our dataset).
\Cref{fig:combo_polygon_centroid_relative_distribution} shows the distribution of centroid positions (relative to the image size).

\begin{figure*}[ht]
\centering
\includegraphics[width=1.0\textwidth]{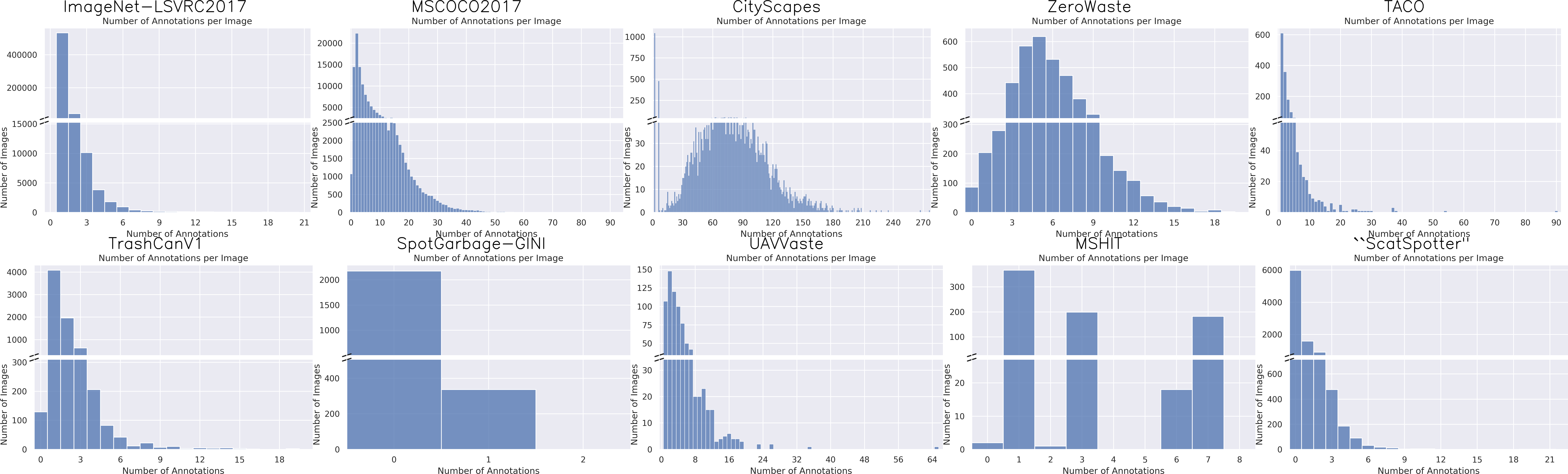}
\caption[]{
    Number of annotations per image in each dataset.
}
\label{fig:combo_anns_per_image_histogram_splity}
\end{figure*}

\begin{figure*}[ht]
\centering
\includegraphics[width=1.0\textwidth]{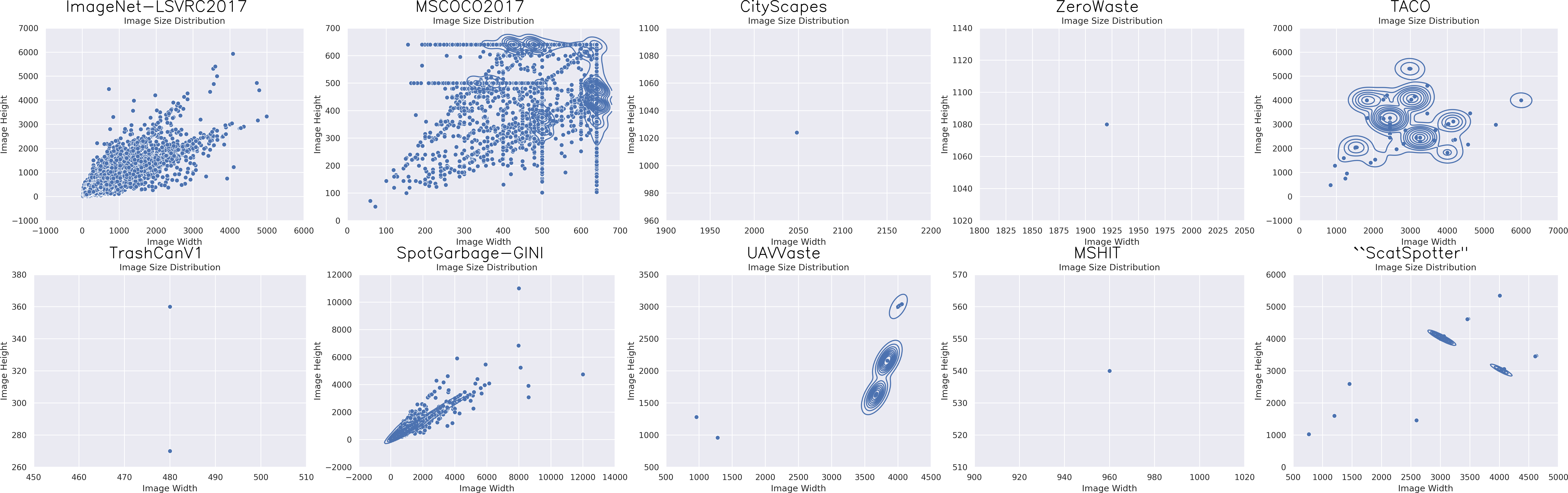}
\caption[]{
    Image size distributions of each dataset. 
    Ours has two primary width/heights.
}
\label{fig:combo_image_size_scatter}
\end{figure*}

\begin{figure*}[ht]
\centering
\includegraphics[width=1.0\textwidth]{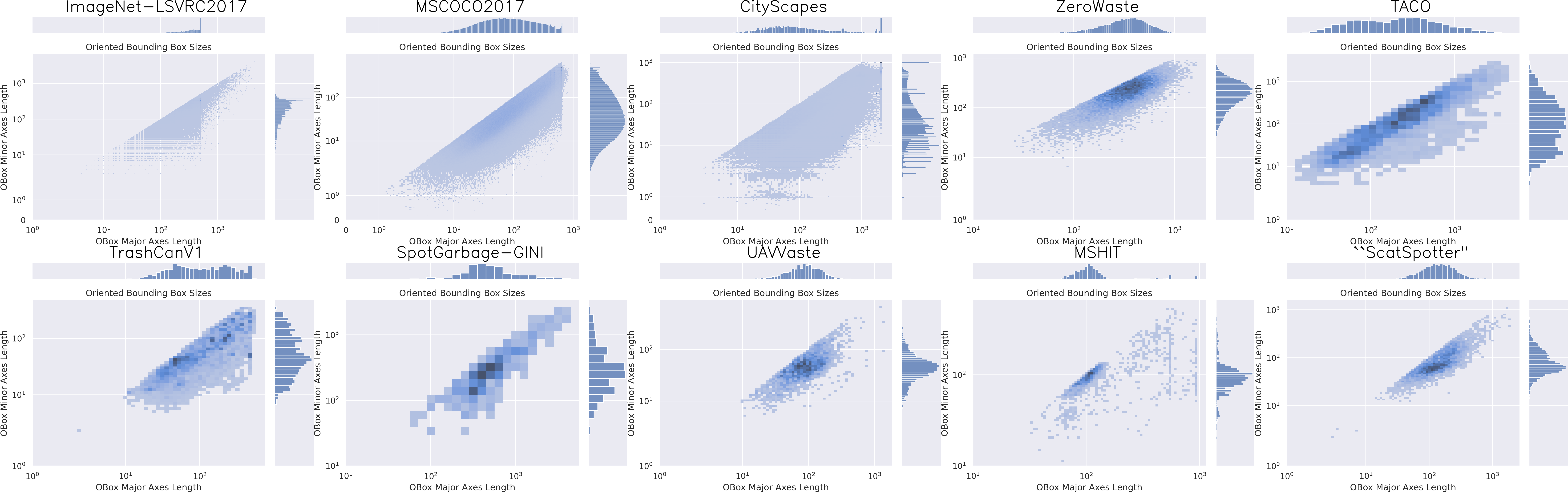}
\caption[]{
    Oriented bounding box size distributions (log10 scale) of each dataset.
}
\label{fig:combo_obox_size_distribution_logscale}
\end{figure*}

\begin{figure*}[ht]
\centering
\includegraphics[width=1.0\textwidth]{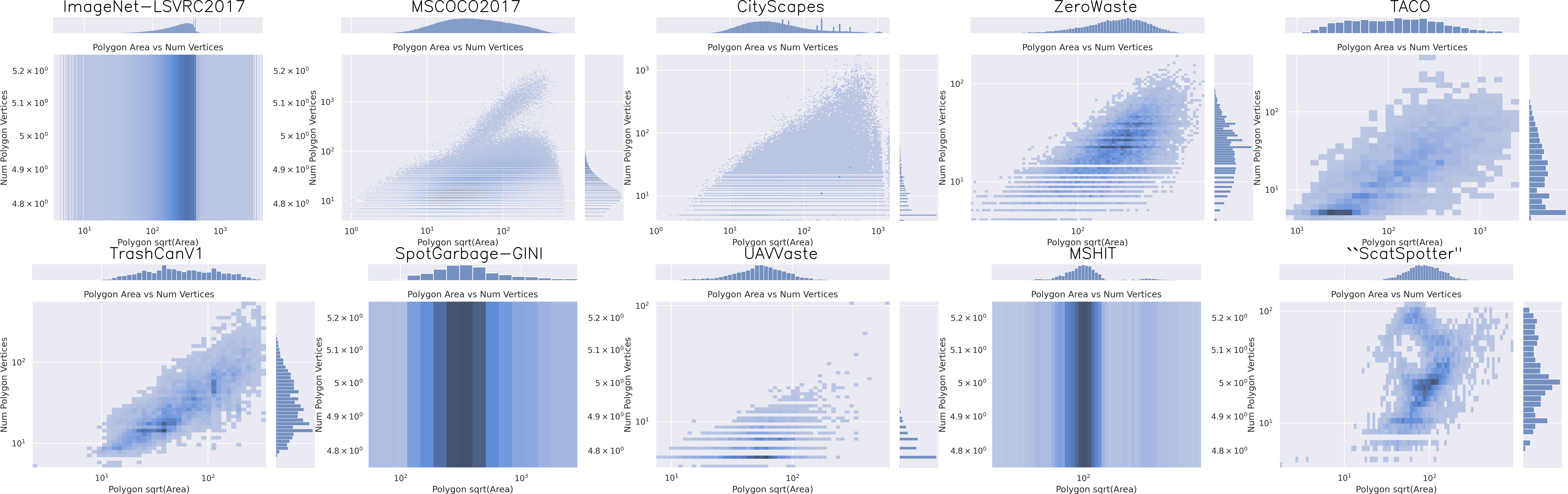}
\caption[]{
    Polygon area versus number of vertices (log10 scale) for each dataset.
    The polygons with more vertices are more likely to be AI generated.
}
\label{fig:combo_polygon_area_vs_num_verts_jointplot}
\end{figure*}

\begin{figure*}[ht]
\centering
\includegraphics[width=1.0\textwidth]{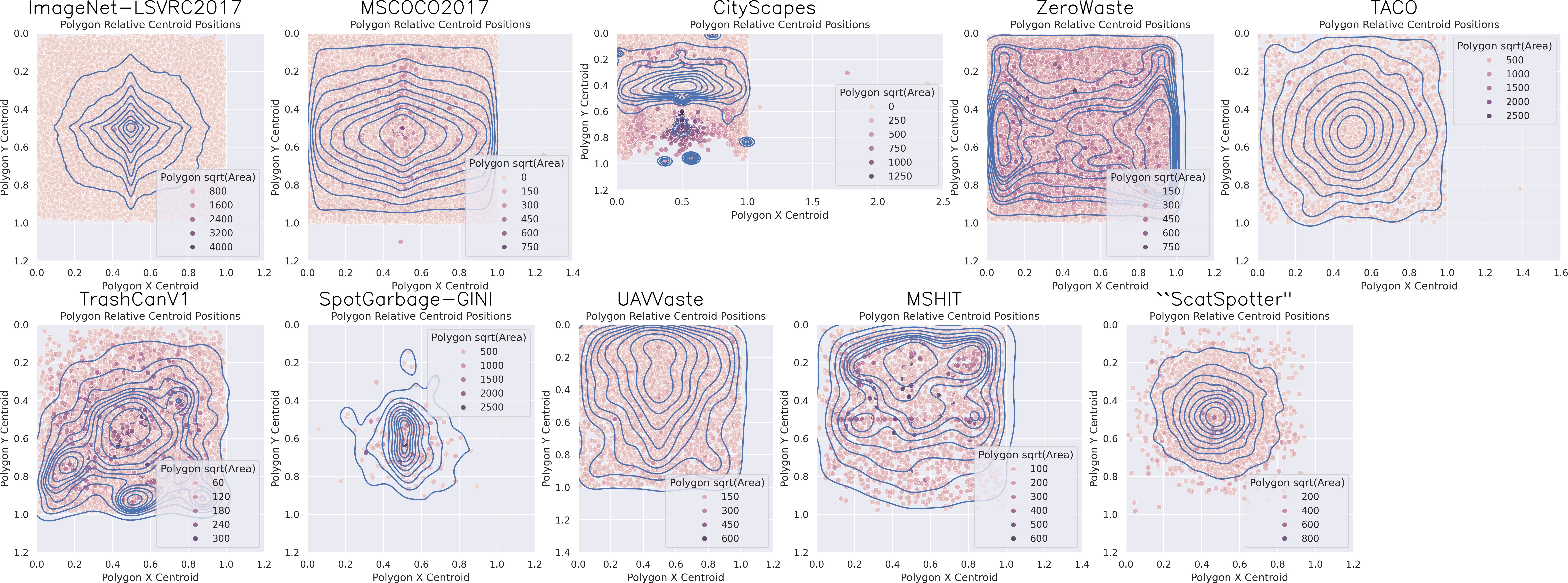}
\caption[]{
    Polygon centroid relative distribution for each dataset. It is interesting
    to note patterns in this data. For instance, the outline of a street can be
    seen in CityScapes. In Zero Waste you can see the conveyor belt. ImageNet
    is more uniform. Ours is Gaussian distributed. 
}
\label{fig:combo_polygon_centroid_relative_distribution}
\end{figure*}


\subsection{Additional Information}
\label{sec:expanded_dataset}

In \Cref{sec:dataset} we provided an overview of several dataset statistics.
In this appendix we expand on that with additional plots.
The distribution of image pixel intensities is illustrated in \Cref{fig:spectra}.
The distribution of images collected over time is shown in \Cref{fig:images_over_time}.
The distribution of annotation location is shown in \Cref{fig:centroid_location_distri} and sizes is shown
  in \Cref{fig:annot_obox_size_dist} and \Cref{fig:annot_area_verts_distri}.

\begin{figure*}[ht]
\centering
\includegraphics[width=0.9\textwidth]{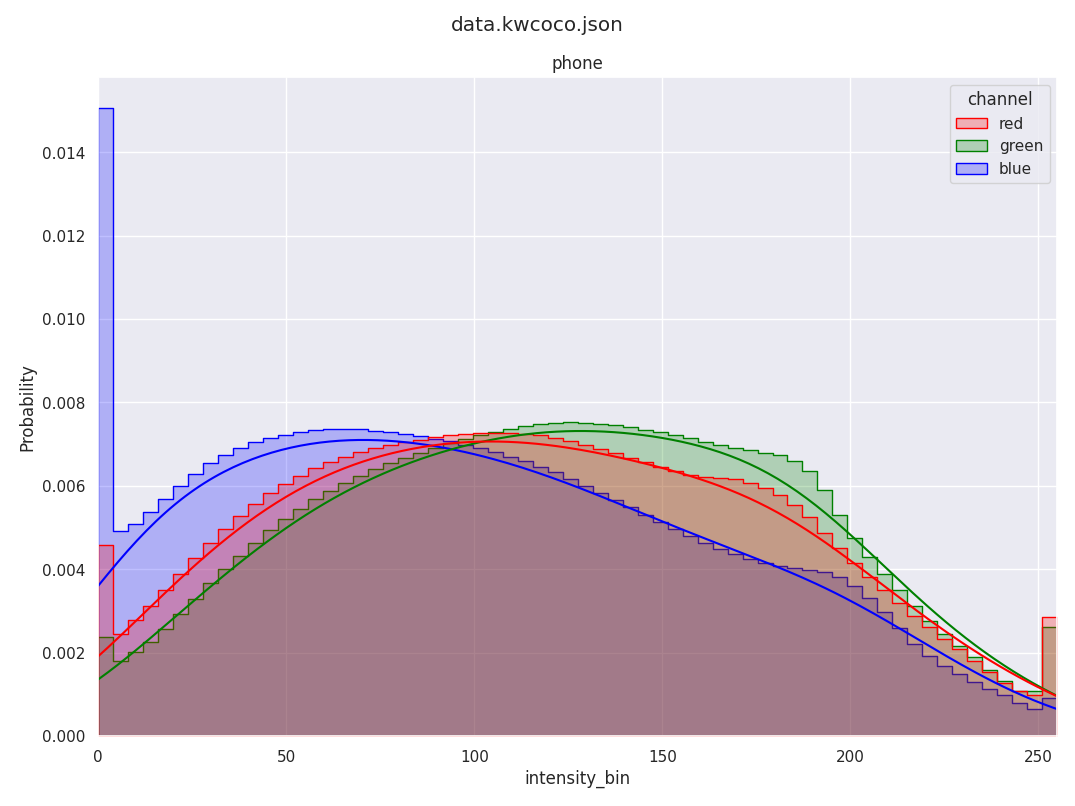}
\caption[]{
    The ``spectra'' or histogram of the pixel intensities in the dataset. 
    The dataset RGB mean/std is $[117, 124, 100], [61, 59, 63]$. 
    High and low saturated values occur, but are included in the stats.
    This was run on the older 2024-07-03 snapshot.
}
\label{fig:spectra}
\end{figure*}

\begin{figure*}[ht]
\centering
\includegraphics[width=0.9\textwidth]{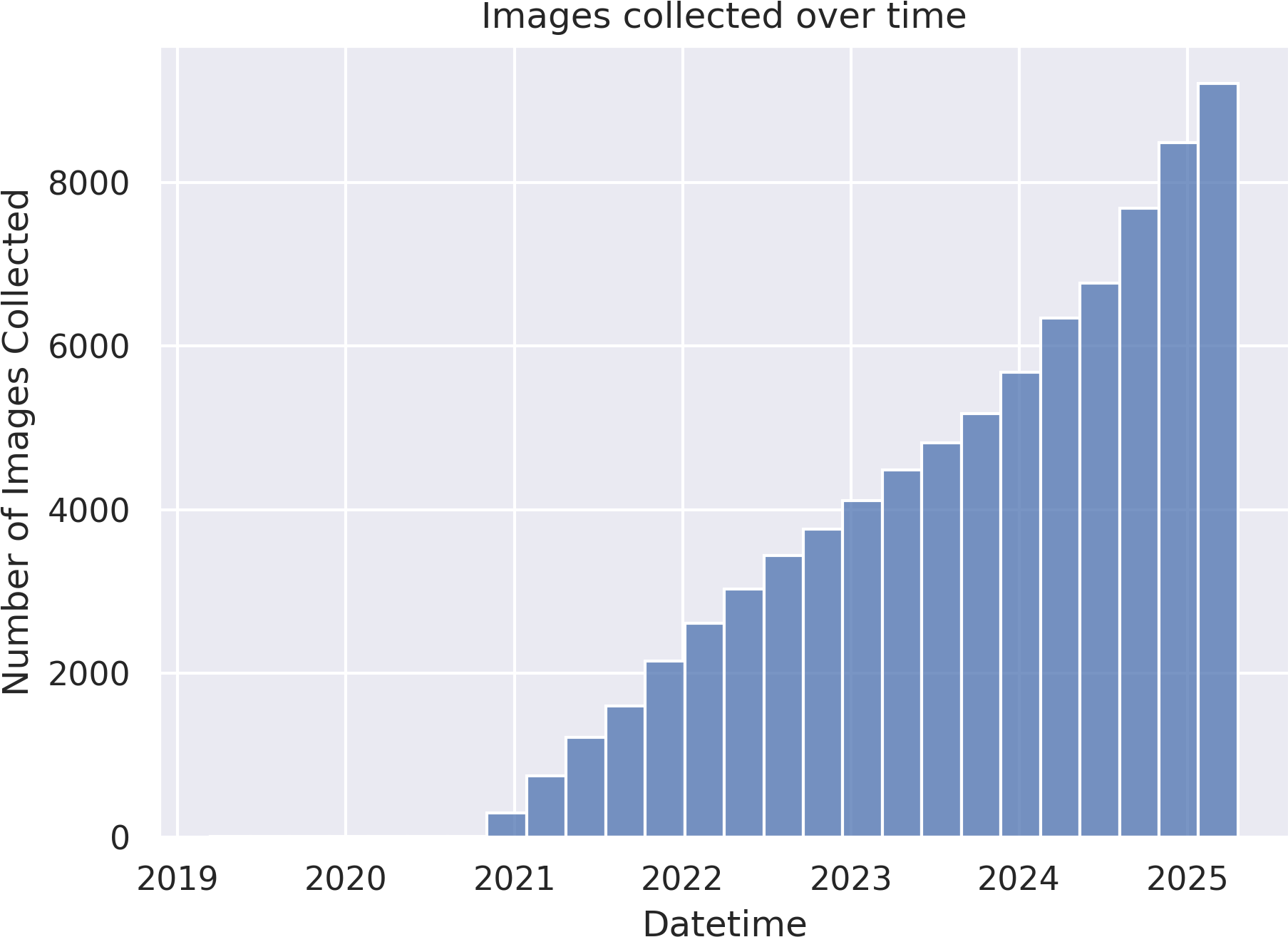}
\caption[]{
    The number of images collected over time.
}
\label{fig:images_over_time}
\end{figure*}

\begin{figure*}[ht]
\centering
\begin{subfigure}[b]{0.4\textwidth}
 \includegraphics[width=\textwidth]{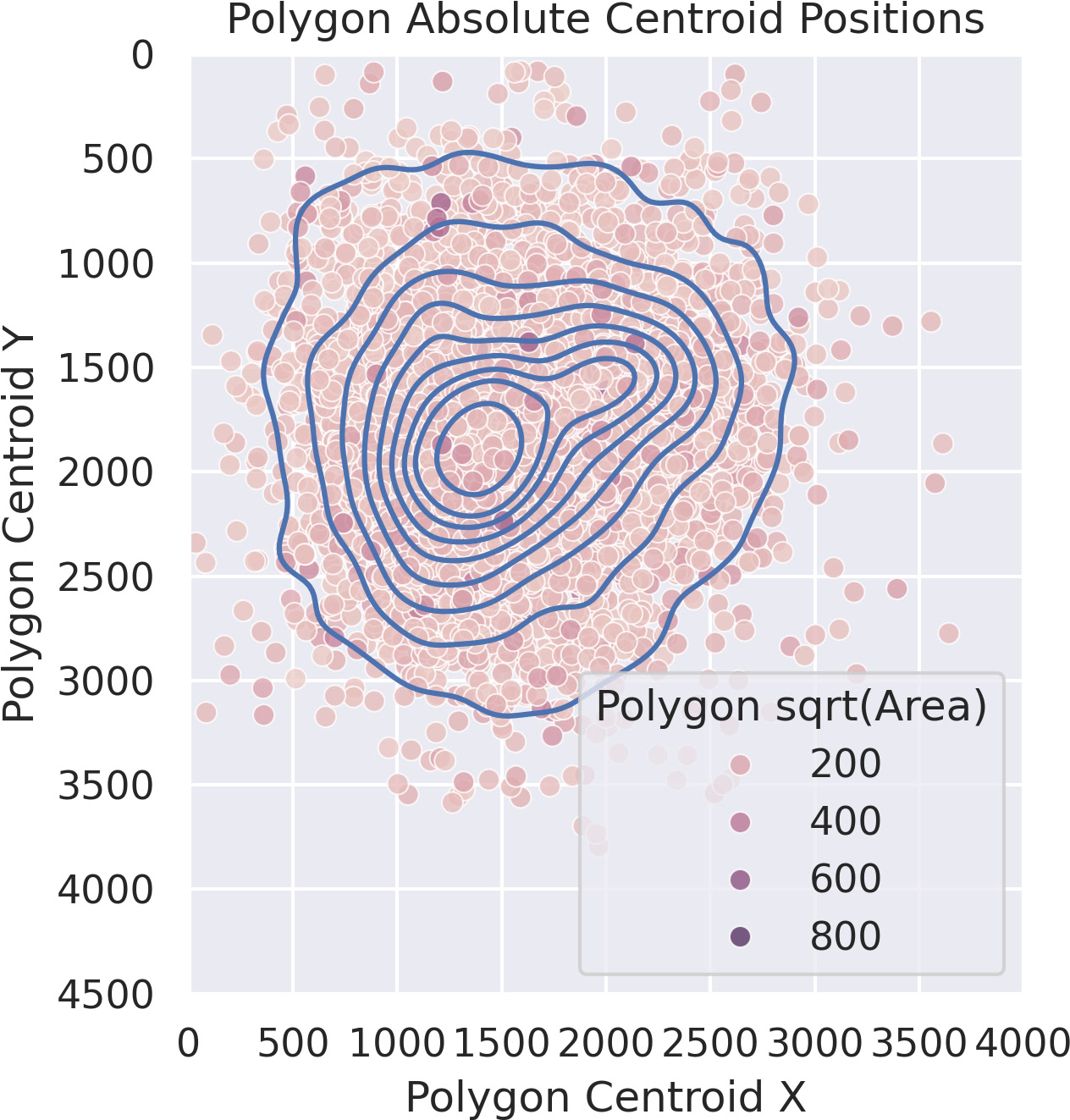}
 \caption{Absolute pixel coordinates.}
 \label{fig:centroid_abs}
\end{subfigure}
\hfill
\begin{subfigure}[b]{0.4\textwidth}
 \includegraphics[width=\textwidth]{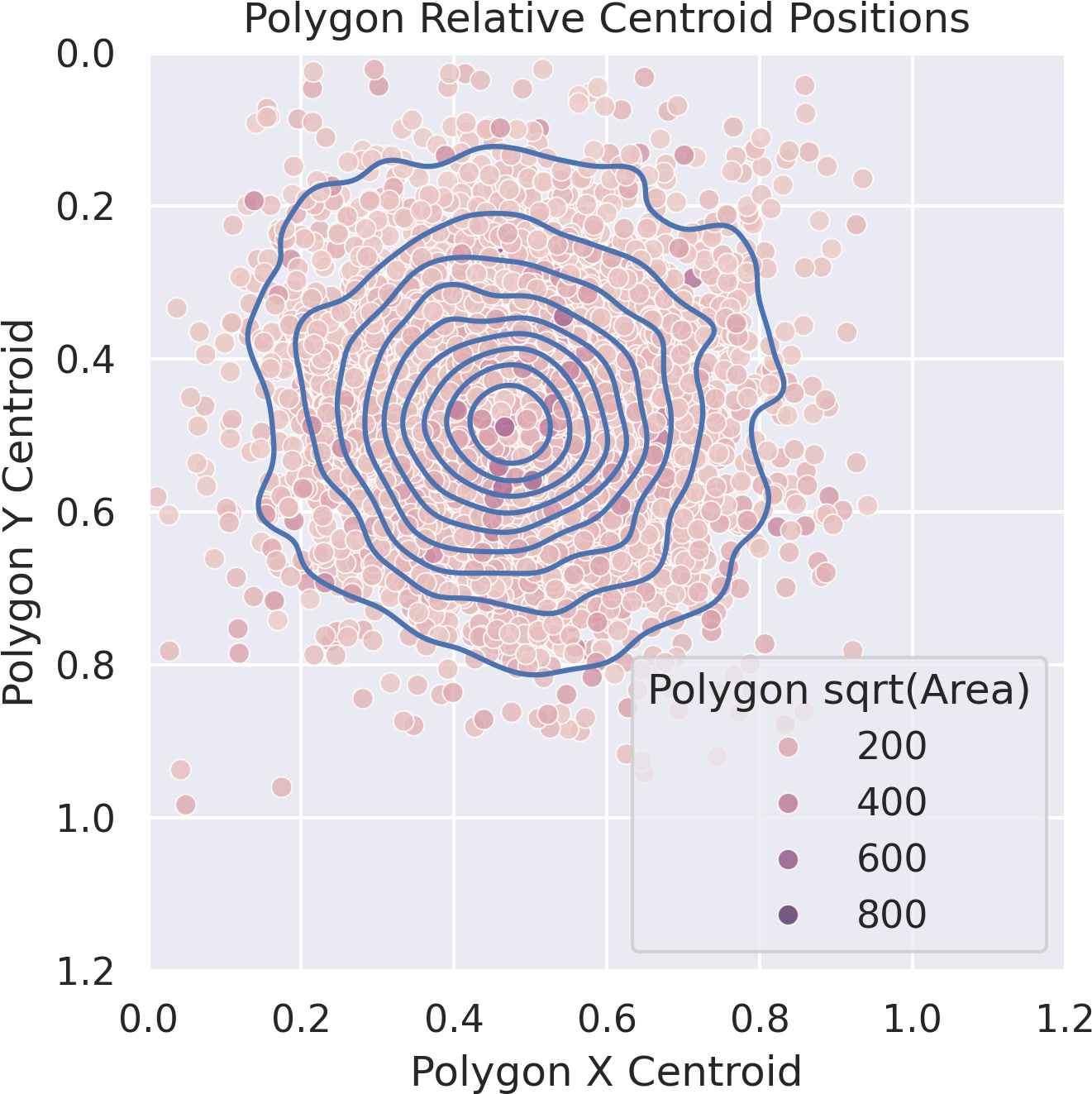}
 \caption{Relative image coordinates.}
 \label{fig:centroid_rel}
\end{subfigure}
\caption{The distribution of annotation centroids in terms of (a) absolute image coordinates and (b) relative image coordinates. The absolute centroid distribution is bimodal because some images are taken in landscape mode and other in portrait mode.}
\label{fig:centroid_location_distri}
\end{figure*}

\begin{figure*}[ht]
\centering
\begin{subfigure}[b]{0.4\textwidth}
  \includegraphics[width=\textwidth]{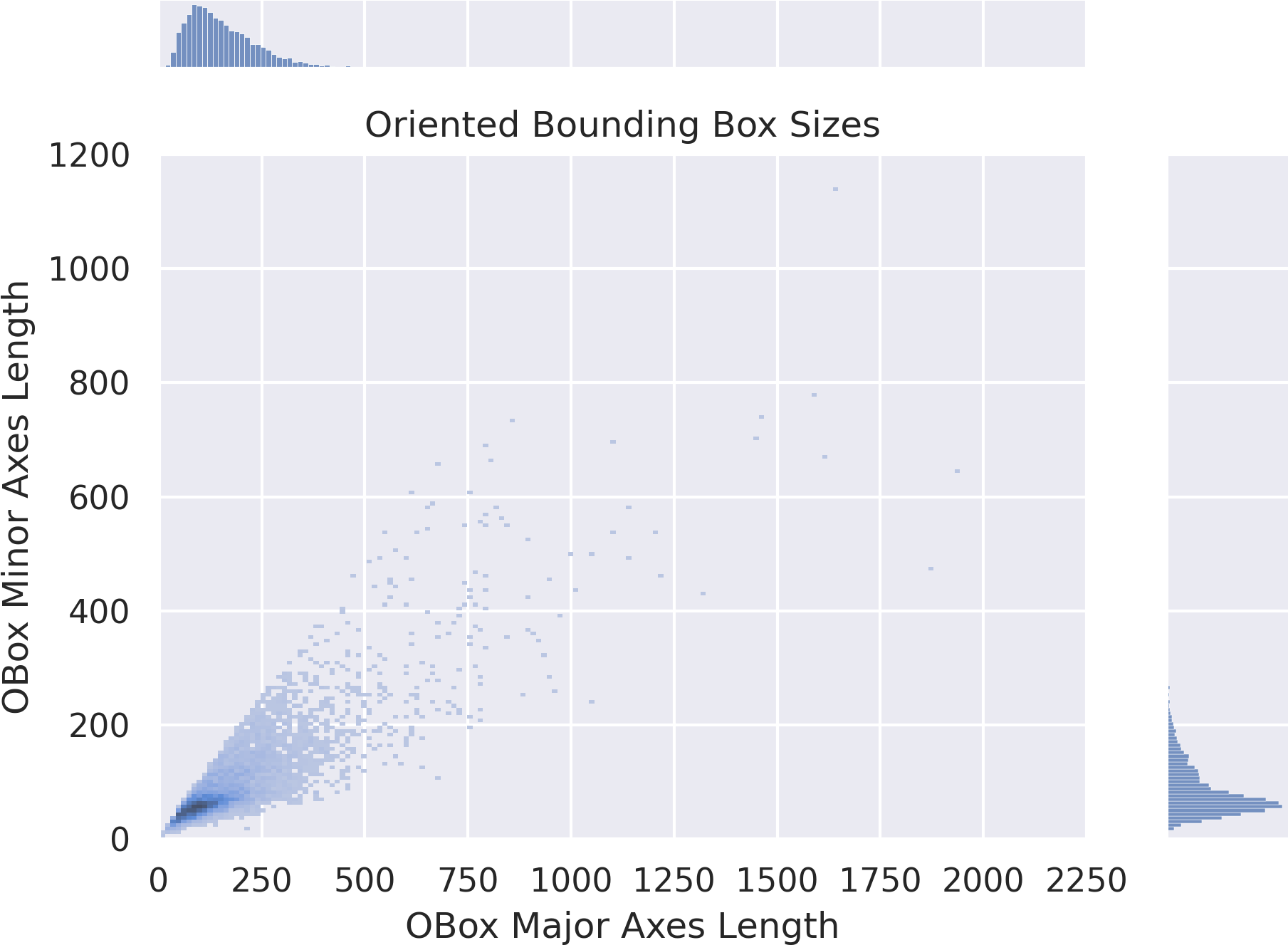}
  \caption{Linear scale.}
  \label{fig:annot_obox_size_dist_linear}
\end{subfigure}
\hfill
\begin{subfigure}[b]{0.4\textwidth}
  \includegraphics[width=\textwidth]{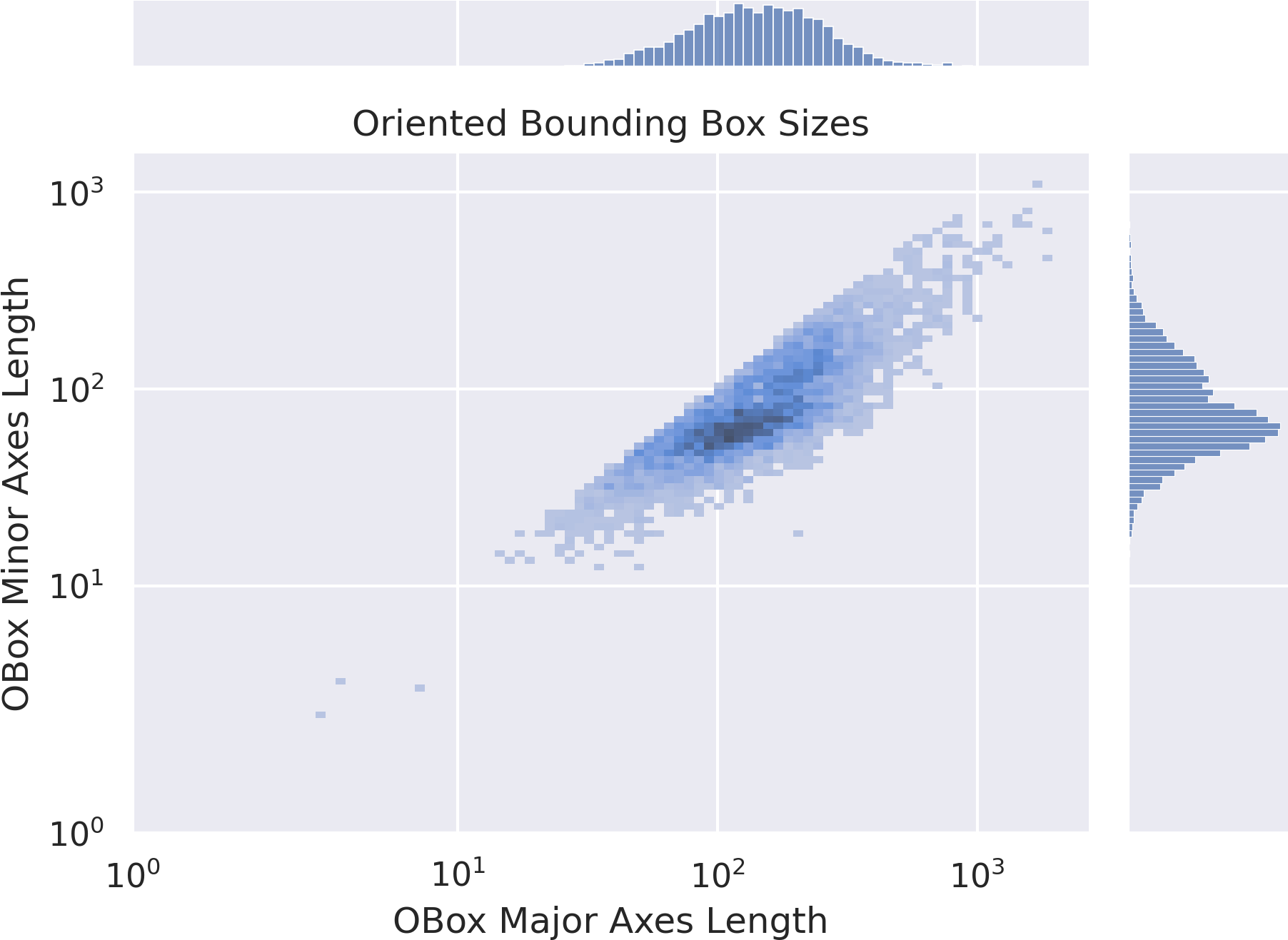}
  \caption{Log10 scale.}
  \label{fig:annot_obox_size_dist_log}
\end{subfigure}
\caption{The distribution of annotation sizes as measured by an oriented bounding box fit to each polygon. (a) shows this plot on a linear scale and (b) show this plot on a log scale.}
\label{fig:annot_obox_size_dist}
\end{figure*}

\begin{figure*}[ht]
\centering
\includegraphics[width=1.0\textwidth]{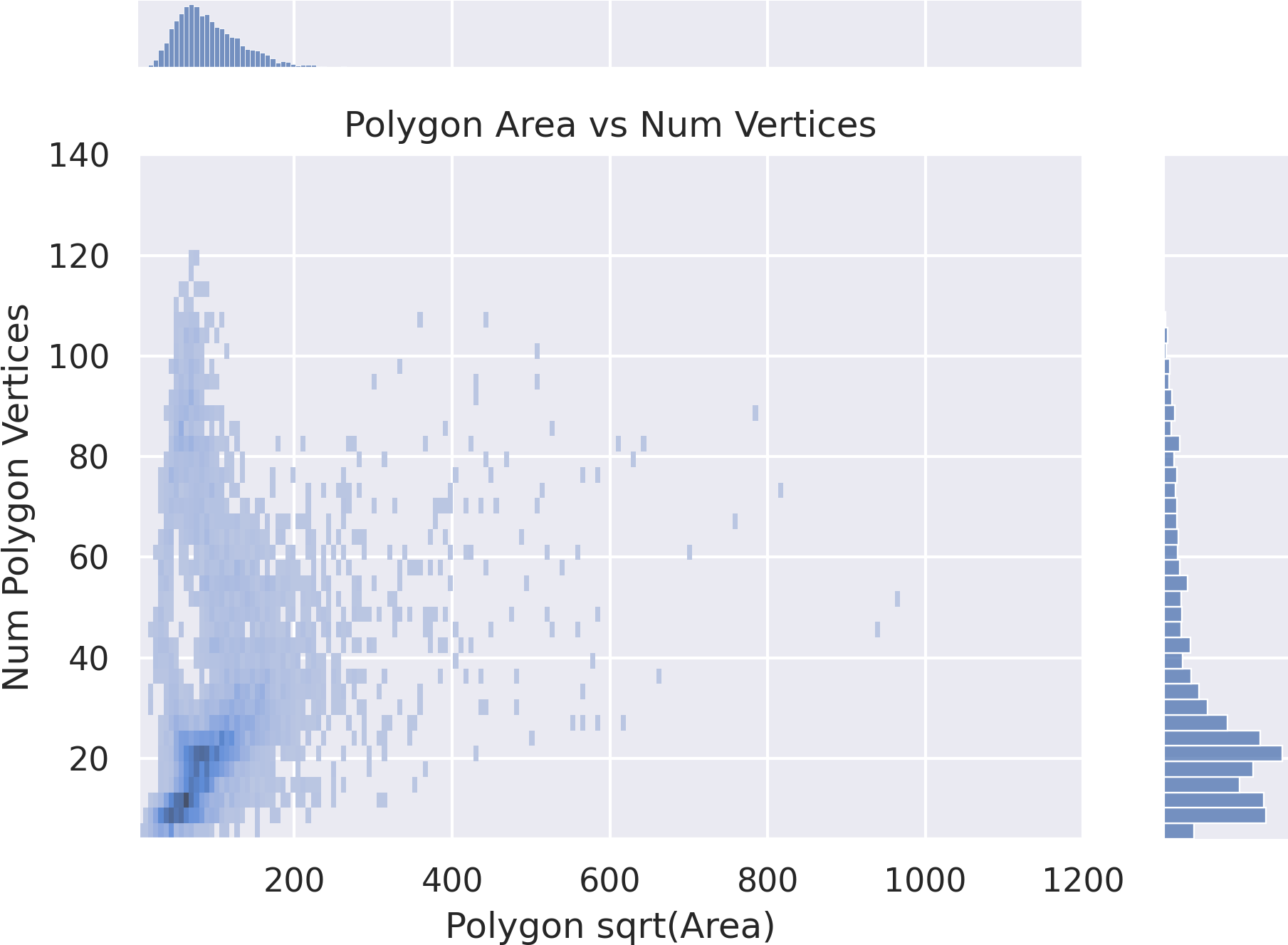}
\caption[]{
    The distribution of polygon areas versus the number of vertices in the polygon boundary.
    The SAM model tends to produce polygons with a higher number of vertices
    than manually drawn ones.  For smaller polygons there are two peaks in the
    number of vertices histograms likely corresponding to pure-manual versus
    AI-assisted annotations.
}
\label{fig:annot_area_verts_distri}
\end{figure*}


\section{Data Distribution \& Transfer}
\label{sec:distribution}

In \Cref{sec:dataset_transfer} we briefly presented a brief set of data distribution experiments.
Here, we provide more background detail, motivation, and discussion.

Empirical evidence suggests that a substantial proportion of scientific studies have low reproducibility
  rates, which has raised concerns across various disciplines \cite{baker_reproducibility_2016}.
Ideally, scientific research should be independently reproducible.
Despite higher success rates in computer science (up to 60\%) compared to other fields, there is still room for improvement
\cite{NEURIPS2019_c429429b, collberg2016repeatability, desai_what_2024}.
Addressing this issue requires not just better experimental documentation but also more reliable and
  accessible data distribution methods.
Specifically, this involves robustly codifying data download and preparation processes.

Centralized data distribution methods allow for codified data access by storing URLs that point to datasets
  within the code, offering fast and direct access.
However, this approach lacks robustness.
It can fail if the provider goes offline, changes the URL, or stops hosting the data.
Additionally, cloud storage can be expensive, and users must trust that the provider delivers the correct
  data --- a risk that can be mitigated by using checksums to verify data integrity.

In contrast, decentralized methods allow users to access data in the same way, even if the organization
  hosting the data changes.
By leveraging content-addressable storage, where the dataset checksum acts as both the key to locate and
  validate the data, these methods ensure data integrity and nearly eliminate the risk of dead URLs, provided
  that at least one peer retains the data.
While decentralized systems face challenges such as longer connection times, increased network overhead, and
  the need for a robust peer network, their ability to ensure data access via a static address
  motivates our investigation

Specifically, we focus on two prominent candidates:
BitTorrent and IPFS.
BitTorrent \cite{cohen_incentives_2003, cohen_bittorrent_2017} is a well known sharing protocol that
  originally relied on centralized trackers and databases of torrent files to connect peers.
While trackers and torrent files are still prominent, torrents can be published to a distributed hash table
  (DHT) using the Kademlia algorithm \cite{maymounkov_kademlia_2002}.
This makes it an strong candidate for a decentralized distribution mechanism.
On the other hand, IPFS (InterPlanetary File System) \cite{benet_ipfs_2014, bieri_overview_2021} is a newer
  tool directly build directly on a DHT.
IPFS has been likened to ``a single BitTorrent swarm, exchanging objects within one Git repository''.
Both IPFS and BitTorrent are content addressable at the dataset level, which makes them both appropriate for
  our use case where we seek a static address that can be used to robustly access data.

It is worth noting that git-based \cite{chacon2014progit} systems like
  HuggingFace~\cite{huggingface_datasets} with large file storage do gain some decentralized
  properties via multiple remotes, but not content identifiers.

For practitioners, key concerns are how quickly and reliably data can be accessed.
By comparing decentralized and centralized mechanisms access times for our dataset, we aim to make
  explicit the tradeoffs between the methods and inform decisions on adopting an approach.



\subsection{Data Distribution Discussion}
\label{sec:datset_discuss}

To assess the effectiveness of each mechanism we programmatically download our 42GB dataset and measure the
  time required to complete the transfer.
Each experiment was run five times, machines we controlled were separated by $\sim\!30$ kilometers with an
  average ping time of 48.48 ms.
For each test, we log transfer start and end times along with notes and code (provided in code repo).

While our measurements provide a reasonable estimate of for access time for each mechanism, there are
  notable limitations in our methodology.
First, different machines and networks have different upload and download speeds, and network congestion is
  variable.
For decentralized methods, we lack an automated mechanism separate peer-connection time and actual download
  time.
Additionally, Girder and HuggingFace required data to be packed into compressed archives, improving transfer
  efficiency due to fewer file boundaries.
In decentralized cases, we provide granular access to each file in the dataset, which avoids an extra
  unpacking step and enables sharing of the same file between different versions of the datasets and simpler
  updates, but decreases transfer efficiency.
Due to this, we provide both a compressed and uncompressed rsync baseline.
Another confounding factor is that with decentralized mechanisms the number of seeders is not controlled
  for.
Subsets of the data have been hosted on IPFS for years, and portions of the dataset may be provided by
  unknown members of the network.
For BitTorrent, our initial transfers only had one seeder, but during our tests other nodes accessed and
  started to provide the data.

Despite significant testing limitations, our measurements quantify the expected data-access time penalty to
  gain the advantages of decentralized mechanisms.
With these limitations acknowledged, we present the transfer times statistics in \Cref{tab:transfertime}.
Alongside these measurements, several observations are worth noting.
Transferring files using IPFS had significantly delayed peer discovery times, and we were only able to
  connect two machines after manually informing them of each other's peer ID.
For BitTorrent, were unable to use the mainline DHT and fell back to using trackers.
We believe these peer discovery issues are because the dataset has a small number of seeders.
To test this, we downloaded other established datasets via IPFS and BitTorrent and found that the peer
  discovery time was almost immediate, suggesting that this becomes less of an issue as a dataset is shared.
However, the inability to quickly find a nearby peer is a major issue for initial or private dataset
  development.

\subsection{Dataset Versions}

An advantage of content identifiers is that they are resistant to link rot as long as at least one peer
  hosts the data, and more importantly, they can never resolve to the wrong content.
This makes them highly attractive for scientific reproducibility.
In this work we relied on two main dataset versions, each specified by a stable content-based identifiers:

\paragraph{Version from 2024-07-03}
\begin{itemize}
  \item IPFS CID: \ipfscid{bafybeiedwp2zvmdyb2c2axrcl455xfbv2mgdbhgkc3dile4dftiimwth2y}
  \item BitTorrent: \magnetlink{ee8d2c87a39ea9bfe48bef7eb4ca12eb68852c49}
\end{itemize}

\paragraph{Version from 2025-04-20}
\begin{itemize}
  \item IPFS CID: \ipfscid{bafybeia2uv3ea3aoz27ytiwbyudrjzblfuen47hm6tyfrjt6dgf6iadta4}
  \item BitTorrent: \magnetlink{27a2512ae93298f75544be6d2d629dfb186f86cf}
\end{itemize}
Note: the hash suffix of the magnet URL can be searched on \url{academictorrents.com}.

At the time of writing, the version of the dataset on HuggingFace is the latest, and we use git tags that
  correspond with the date of release and the IPFS CID to help identify dataset versions.
However, unlike the decentralized methods, these are not guaranteed to point to the expected version of the
  dataset.
At the time of writing the HuggingFace URL is:
\url{https://huggingface.co/datasets/\redact{erotemic}/scatspotter} and the Girder URL is:
\url{https://data.\redact{kitware}.com/?#user/598a19658d777f7d33e9c18b/folder/66b6bc7ef87a980650f41f98}.


\section{Model \& Training Details}
\label{sec:experiment_details}

In \Cref{sec:models} we provided our main results.
However, a key limitation of these results is the imbalance between model types, with 42 of 47 trained
  models being VIT-ssegs, 2 MaskRCNN models, 2 YOLO models, and 1 tuned GroundingDino model.
Future work could further optimize MaskRCNN, Grounding~DINO, YOLO, and other models to improve both
  performance and comparability, but these results are enough to establish a useful baseline.

For non-VIT models we adhered as closely as possible to the default parameters of their respective
  frameworks, applying changes needed to support our generalized KWCoco format and to fit on a single GPU.
For complete details, we provide links to the training and evaluation scripts for each model family (see
  below).
A docker image with dependencies pre-installed is also available:
\dockerimage
  {https://hub.docker.com/layers/\redact{erotemic}/shitspotter/latest/images/sha256-aec306e515a5c8bef162c872c96b6a82ff3f4798f4b796f1431ce8f1f6288747}
  {\redact{erotemic}/shitspotter:latest}.

\paragraph{Experiment scripts (by task).}

\begin{itemize}
  \item \textbf{Grounding~DINO:} \repolink{experiments/grounding-dino-experiments}{./experiments/grounding-dino-experiments}

  \item \textbf{YOLO-v9:} \repolink{experiments/yolo-experiments}{./experiments/yolo-experiments}

  \item \textbf{MaskRCNN:} \repolink{experiments/detectron2-experiments}{./experiments/detectron2-experiments}

  \item \textbf{VIT-sseg:} \repolink{experiments/geowatch-experiments}{./experiments/geowatch-experiments}
\end{itemize}

\subsection{Grounding Dino}

GroundingDINO was evaluated in two modes.
The zero-shot setting used the \hflink{IDEA-Research/grounding-dino-tiny} model from HuggingFace, applied
  directly to our validation and test splits with a fixed set of 10 prompts.
The tuned variant used the community \ghlink{longzw1997/Open-GroundingDino} implementation, initialized from
  \DINOPretrained{} with a BERT text encoder on a single GPU.
Training data was converted to ODVG JSONL format, and the label set reduced to two classes (``poop'' and
  ``unknown'').

The preprocessing pipeline resized the shorter side of each image to 800 pixels while maintaining aspect
  ratio, corresponding to a median scale factor of $\sim$0.26 for our dataset.

In the zero-shot evaluation, we tested ten prompts and selected ``animalfeces'' based on the highest
  validation Box-AP.
\Cref{tab:prompt_variations} shows the full ablation, illustrating that prompt choice strongly affects
  performance.
Prompt choice has a large effect, and the best prompt differs between validation and test splits, but
  overall zero-shot results remain low.


\newcommand{\CaptionPrompt}{
\caption{Zero-shot detection results with varied prompts. The chosen prompt has a significant impact on scores, and the best prompt is different between validation and test datasets, but overall zero-shot results are all low scoring. Because this is a zero-shot setting, the validation set can be compared to the test set. Interestingly, the validation scores significantly lower than the test scores indicating a greater degree of difficulty. }
\label{tab:prompt_variations}
}

\begin{table*}[t]
\ifwacv \else \CaptionPrompt \fi
\centering
\begin{tabular}{lllllllll}
\toprule
\multicolumn{1}{l}{} & \multicolumn{4}{c}{Validation (n=691)} & \multicolumn{4}{c}{Test (n=121)} \\
 Prompt      & \makecell{AP\\Box}   & \makecell{AUC\\Box}   & \makecell{F1\\Box}   & \makecell{TPR\\Box}   & \makecell{AP\\Box}   & \makecell{AUC\\Box}   & \makecell{F1\\Box}   & \makecell{TPR\\Box}   \\
\midrule
 stool       & 0.01                 & 0.03                  & 0.07                 & 0.07                  & 0.05                 & 0.08                  & 0.18                 & 0.13                  \\
 droppings   & 0.02                 & 0.10                  & 0.14                 & 0.23                  & 0.08                 & 0.14                  & 0.27                 & 0.30                  \\
 petwaste    & 0.04                 & 0.14                  & 0.15                 & 0.25                  & 0.20                 & 0.25                  & 0.35                 & 0.34                  \\
 poop        & 0.04                 & 0.10                  & 0.17                 & 0.16                  & 0.17                 & 0.18                  & 0.31                 & 0.26                  \\
 dogpoop     & 0.05                 & 0.16                  & 0.17                 & 0.20                  & 0.24                 & 0.28                  & 0.38                 & 0.39                  \\
 caninefeces & 0.05                 & 0.16                  & 0.18                 & 0.29                  & 0.17                 & 0.24                  & 0.37                 & 0.39                  \\
 turd        & 0.05                 & 0.18                  & 0.18                 & 0.22                  & \textbf{0.27}        & \textbf{0.32}         & 0.39                 & 0.35                  \\
 feces       & 0.06                 & 0.21                  & 0.18                 & 0.27                  & 0.16                 & 0.26                  & 0.32                 & 0.39                  \\
 excrement   & 0.07                 & \textbf{0.22}         & 0.20                 & 0.28                  & 0.25                 & 0.31                  & 0.39                 & \textbf{0.42}         \\
 dogfeces    & 0.07                 & 0.21                  & \textbf{0.20}        & \textbf{0.31}         & 0.23                 & 0.29                  & \textbf{0.40}        & 0.38                  \\
 animalfeces & \textbf{0.08}        & 0.21                  & 0.20                 & 0.25                  & 0.23                 & 0.30                  & 0.39                 & 0.38                  \\
\bottomrule
\end{tabular}
\ifwacv \CaptionPrompt \fi
\end{table*}

\subsection{YOLO-v9}

YOLO-v9 experiments were based on the community \ghlink{WongKinYiu/yolov9} implementation.
We trained both pretrained (ImageNet-initialized) and from-scratch variants, using our fork adapted for
  KWCoco input.

All images were resized to 640$\times$640, corresponding to a median scale factor of $\sim$0.16.
Training used a batch size of 16 with batch accumulation set to 50, for an effective batch size of 800.
Optimization used AdamW with learning rate $3\times10^{-4}$ and weight decay of 0.01.
The pretrained runs started from \YOLOPretrained{}.

\subsection{MaskRCNN}

MaskRCNN experiments were run using a Detectron2 fork with KWCoco support.
Both pretrained and from-scratch models used the standard \texttt{R\_50\_FPN\_3x.yaml} configuration,
  differing only in initialization:
pretrained models used \MaskRCNNPretrained{}, while the from-scratch models started randomly.

To fit training on a single GPU, we reduced the learning rate to $2.5\times10^{-4}$, set the batch size to
  2, and trained for a maximum of 120{,}000 iterations.
The maximum image dimension was capped at 1024, giving a median scale factor of $\sim$0.25 for our
  dataset.

\subsection{VIT-sseg}
\label{sec:vit_models}

This subsection provides additional details on VIT-sseg models, which were the first architecture we
  explored for this problem and therefore have the most extensive analysis compared to other networks.

To train VIT-sseg models we use the training, prediction, and evaluation system presented in
  \cite{Greenwell_2024_WACV, crall_geowatch_2024}, which utilizes polygon annotations to train a pixelwise
  binary segmentation model.

In all experiments, we use half-resolution images, which means most images have an effective width $\times$
  height of 2,016 $\times$ 1,512.
We employ a spatial window size of 416 $\times$ 416 for network inputs, which means that multiple windows
  are needed to predict on entire images.
During prediction, we apply a window overlap of 0.3 with feathered stitching to prevent boundary artifacts.

To address the class imbalance in our dataset (where positives are patches containing annotations and
  negatives contain no annotations), we adopt a balanced sampling strategy.
Each ``epoch'' consists of randomly sampling 32,768 patches from the dataset with replacement, ensuring
  roughly equal numbers of positive and negative samples.
We train each network for 163,840 gradient steps.
For data augmentation we use random crops and flips.

Our baseline architecture is a variant \cite{bertasius2021space,Greenwell_2024_WACV} of a vision-transformer
  \cite{dosovitskiy_image_2021}.
The model is a 12-layer encoder backbone with 384 channels and 8 attention heads that feeds into a 4-layer
  MLP segmentation head.
It has 25,543,369 parameters and a size of 114.19 MB on disk.
At predict time it uses 1.96GB of GPU RAM.

We compute loss pixelwise using Focal Loss \cite{ross2017focal} with a small downweighting of pixels towards
  the edge of the window.
Our optimizer is AdamW \cite{loshchilov_decoupled_2018}, and we experiment with varying learning rate,
  weight decay, and perturb-scale (implementing the shrink perturb trick~\cite{ash_warm_starting_2020,dohare_loss_2023}).
We employ a OneCycle learning rate scheduler \cite{smith2019super} with a cosine annealing strategy and
  starting fraction of 0.3.
Our effective batch size is 24 with a real batch size of 2 and 12 accumulate gradient steps.
This setup consumes approximately 20 GB of GPU RAM during training.

\subsubsection{VIT-sseg Model Experiments}

To establish a baseline, we evaluated 35 training runs where we varied input resolutions, window sizes,
  model depth, and other parameters.
Although this initial search was somewhat ad-hoc, it provided insights into the optimal configuration for
  our model.
Building on the best hyperparameters from this search, we performed a sweep over 7 combinations of learning
  rate, weight decay, and perturb scale (i.e., shrink and perturb
  \cite{ash_warm_starting_2020,dohare_loss_2023}).
Scripts used to reproduce these experiments, as well as a log of the ad-hoc experiments, are available in
  the code repository.
Additionally, trained models are packaged and distributed with information about their training
  configuration.

Note:
the test dataset used in this appendix section is an older 30 image version with suffix {\tt d8988f8c},
  which is a subset of the more recent 121 image test set used in the main paper.

\newcommand{\VitResultCaption}{
\caption{
Results for the best-performing models on the validation set across 7 hyperparameter configurations.
The table provides detailed information about each configuration, including:
1) Configuration name (first column): a unique code identifying each training run used in the score scatter and box plots.
2) Varied hyperparameters (next three columns): specific values for learning rate, weight decay, and perturb scale that were used in each run.
3) Validation set performance (AP and AUC scores): metrics evaluating the model's performance on the validation set.
4) Test set performance (AP and AUC scores): metrics evaluating the model's performance on the test set using the same validation-maximizing models.
Note that the top AP score over all models on the test set was 0.65, but it did not correspond to one of these validation runs used for model selection.
Qualitative examples illustrating the performance of the top-scoring validation model listed here are provided in \cref{fig:test_heatmaps_with_best_vali_model}.
}
\label{tab:parameters_and_results}
}

\begin{table*}[t]
\ifwacv \else \VitResultCaption \fi
\centering
\begin{tabular}{llllllll}
\toprule
            \multicolumn{4}{l}{} & \multicolumn{2}{r}{Validation (n=691)} & \multicolumn{2}{r}{Test (n=30)} \\
Config Name  &   LR & Weight Decay & Perterb Scale & \makecell{AP\\Pixel} & \makecell{AUC\\Pixel} &  \makecell{AP\\Pixel} &  \makecell{AUC\\Pixel} \\
\midrule
        \textcolor[HTML]{623682}{D05} & 1e-4 &   1e-6 &  3e-6 & \textbf{0.7802} & \textbf{0.9943} &          0.5051 &          0.9125 \\
        \textcolor[HTML]{df8020}{D03} & 1e-4 &   1e-5 &  3e-7 &          0.7758 &          0.9707 &          0.4346 &          0.8576 \\
        \textcolor[HTML]{87b787}{D04} & 1e-4 &   1e-7 &  3e-7 &          0.7725 &          0.9818 &          0.4652 &          0.7965 \\
        \textcolor[HTML]{207fdf}{D02} & 1e-4 &   1e-6 &  3e-7 &          0.7621 &          0.9893 & \textbf{0.5167} & \textbf{0.9252} \\
        \textcolor[HTML]{20df20}{D00} & 3e-4 &   3e-6 &  9e-7 &          0.7571 &          0.9737 &          0.4210 &          0.7766 \\
        \textcolor[HTML]{df20df}{D01} & 1e-3 &   1e-5 &  3e-6 &          0.7070 &          0.9913 &          0.4607 &          0.9062 \\
        \textcolor[HTML]{b00403}{D06} & 1e-4 &   1e-6 &  3e-8 &          0.6800 &          0.9773 &          0.4137 &          0.8157 \\
        
\bottomrule
\end{tabular}
\ifwacv \VitResultCaption \fi
\end{table*}

For each of the 7 hyperparameter combinations, we trained the model for 163,840 optimizer steps using a
  batch size of 24.
We defined an ``epoch'' as 1,365 steps, at which point we saved a checkpoint, evaluated validation loss, and
  adjusted learning rates.
To conserve disk space, we retained only the top 5 lowest-validation-loss checkpoints (although training
  crashes and restarts sometimes resulted in additional checkpoints, which are included in our evaluation).

Using the top-checkpoints, we predicted heatmaps for each image in the validation set.
We then performed binary classification on each pixel (poop-vs-background) using a threshold.
Next, we rasterized the truth polygons.
The corresponding truth and predicted pixels were accumulated into a confusion matrix, allowing us to
  compute standard metrics such as precision, recall, false positive rate, etc.
\cite{powers_evaluation_2011} for the specific threshold.
By sweeping a range of thresholds, we calculated the average precision (AP) and the area under the ROC curve
  (AUC).
We computed all metrics using scikit-learn \cite{scikit-learn}.
Due to the high number of true negative pixels, we preferred AP as the primary measure of model quality.
  
The details of the top model for each run, along with relevant hyperparameters, are presented in
  \Cref{tab:parameters_and_results}.
This table also includes the results on the small, held out, test set for the top model.

The results show strong performance on the validation set, with a maximum AP of $0.78$.
However, while the test AP for this model is good, it is significantly lower at $0.51$.
To investigate this discrepancy, we turned to qualitative analysis.

Qualitative results for the test, validation, and training sets are presented in
  \cref{fig:test_heatmaps_with_best_vali_model}.
These examples illustrate both success and failure cases.
The test and validation sets show clear responses to objects of interest, but the test set contains images
  of close-up and partially deteriorated poops.
This suggests a bias in the dataset towards ``fresh'' poops taken from some distance.

Notably, the much larger training set also contains errors, indicating more information can be extracted
  from this dataset using hard-negative mining.
There are clear difficult cases caused by sticks, leaves, pine cones, and dark areas on snow.
We note that while compiling these results, we checked over 1000 images and discovered 14 cases where an
  object failed to be annotated, and it is likely that more are missed, but we believe these cases are rare.

Although focal loss was used, the current learning curriculum is likely under-weighting smaller distant
  objects.
Our pixelwise evaluation metric is biased against this, which is a current limitation of our approach.
Future work evaluating this dataset on an object-detection level can remedy this.

\begin{figure*}[ht]
\centering
\includegraphics[width=1.0\textwidth]{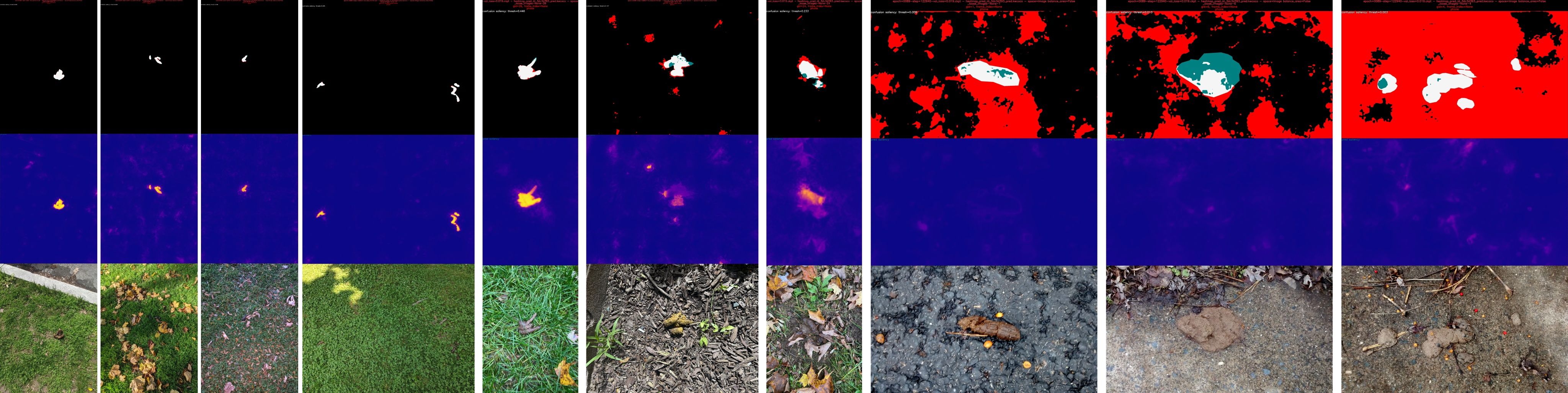}%
\hfill
(a) Test set.
\includegraphics[width=1.0\textwidth]{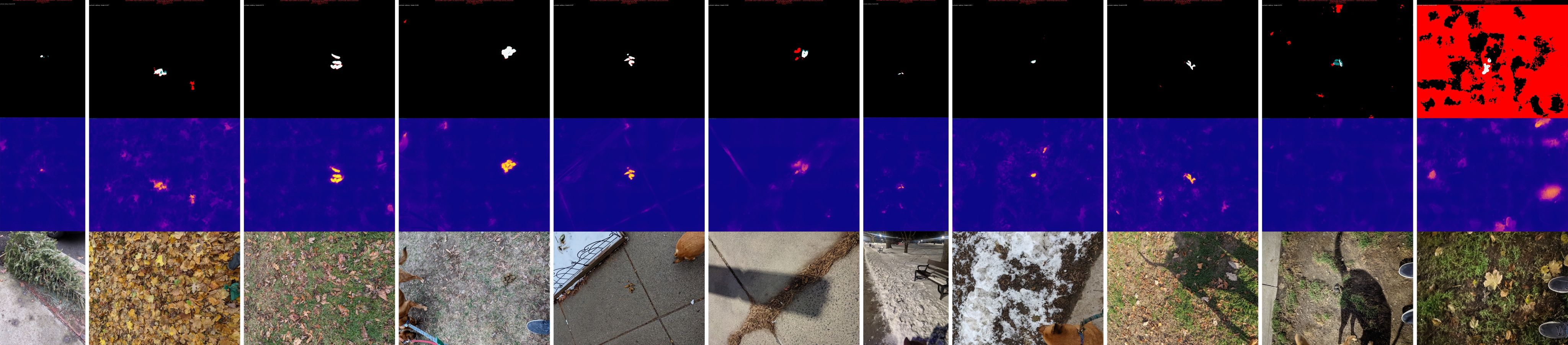}%
\hfill
(b) Validation set.
\includegraphics[width=1.0\textwidth]{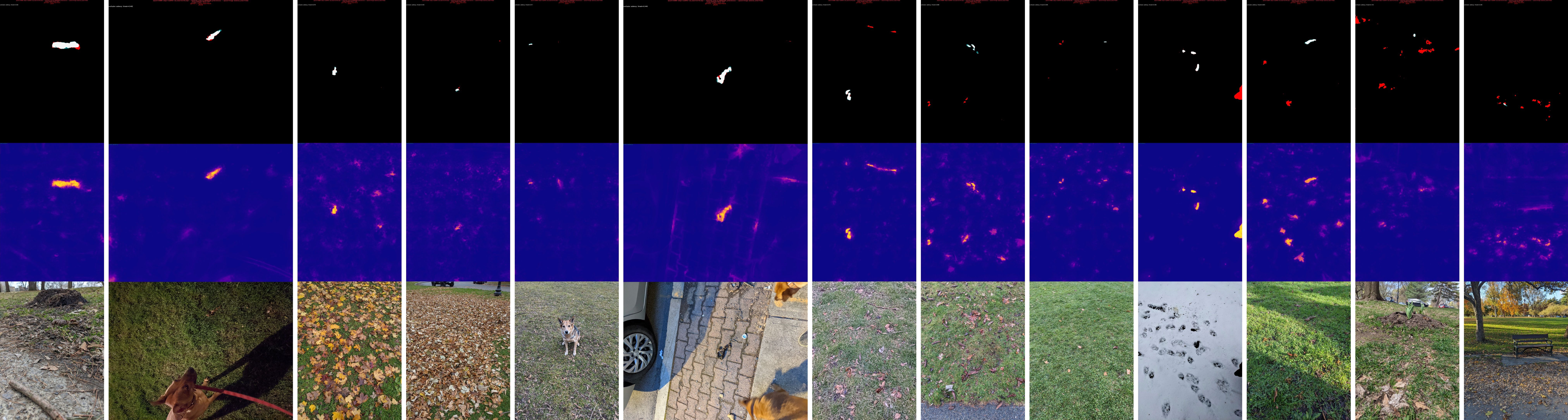}%
\hfill
(c) Training set.
\caption[]{
    Qualitative results using the top-performing model on the validation set, applied to a selection of images
      from the (a) test, (b) validation, and (c) training sets.
    Success cases are presented on the left, with failure cases increasing towards the right.
    Each figure is organized into three rows:
    Top row:
    Binarized classification map, where true positive pixels are shown in white, false positives in red, false
      negatives in teal, and true negatives in black.
    The threshold for binarization was chosen to maximize the F1 score for each image, showcasing the best
      possible classification of the heatmap.
    Middle row:
    The predicted heatmap, illustrating the model's output before binarization.
    Bottom row:
    The input image, providing context for the prediction.
    The majority of images in the test set exhibit qualitatively good results.
    Failure cases tend to occur with close-up images of older, sometimes partially deteriorated poops.
    These examples were manually selected and ordered to demonstrate dataset
    diversity in addition to representative results.
}
\label{fig:test_heatmaps_with_best_vali_model}
\end{figure*}

In \Cref{tab:parameters_and_results} we only presented the top results.
Here we've plotted the AP and AUC on the validation set for the top 5 AP-maximizing results from each of the
  7 training runs.
We also created a box-and-whisker plot for these top 5 results, which serves to assign a color and label to
  each training run.
These plots are shown in \Cref{fig:apauc_scatter}.

\begin{figure}[ht]
\centering
\begin{subfigure}[b]{0.4\textwidth}
 \includegraphics[width=\textwidth]{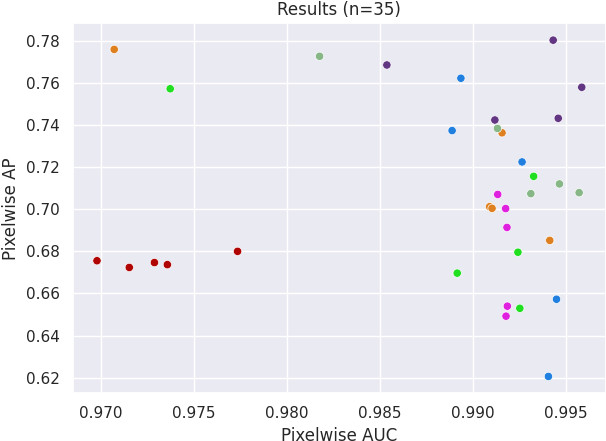}
 \caption{AP and AUC of 35 checkpoints.}
 \label{fig:apauc_scatter_a}
\end{subfigure}
\hfill
\begin{subfigure}[b]{0.4\textwidth}
 \includegraphics[width=\textwidth]{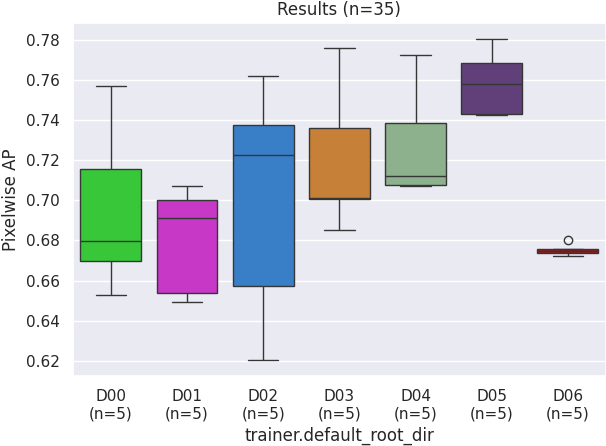}
 \caption{AP of 35 checkpoints.}
 \label{fig:apauc_scatter_b}
\end{subfigure}
\caption{
    (a) Scatterplot of pixelwise average precision (AP) and Area Under the ROC curve (AUC) for the top
      5 checkpoints on the validation set.
    Points of the same color represent checkpoints from the same training run, which used identical
      hyperparameters.
    (b) Box-and-whisker plot the AP values across the top 5 checkpoints evaluated on
      the validation set.
    For each run, corresponding varied hyperparameters and maximum APs are given in
      \Cref{tab:parameters_and_results}.
}
\label{fig:apauc_scatter}
\end{figure}

\newcommand{\VitResourceCaption}{
\caption[]{
Resources used for training, prediction, and evaluation.
The "node" column is the pipeline stage:
"train" for training, "pred" for heatmap prediction, and "eval" for pixelwise heatmap evaluation.
The "resource" column lists the resource type: time, energy, or emissions.
The "total" and "$\mu$" columns show the total and average consumptions, and the "n" column indicates the
  frequency of each stage (e.g., across different hyperparameters).
Train rows marked with an asterisk (*) are based on indirect measurements.
}
\label{tab:resources}
}

\begin{table*}[ht]
\ifwacv \else \VitResourceCaption \fi
  \centering
  \begin{subtable}[b]{\textwidth} 
    \caption{Presented VIT experiment resources.}
    \centering
    \begin{tabular}{llllr}
    \toprule
            Node & Resource    &           Total  &          $\mu$ &  n \\
    \midrule
    eval        &        time  & 14.24 hours      & 0.41 hours     &   35 \\
    \rule{0pt}{2ex}%
    pred        &        time  & 11.97 hours      & 0.34 hours     &   35 \\
    pred        &      energy  &  8.76 kWh        & 0.25 kWh       &   35 \\
    pred        &   emissions  &  1.84 \cotwo kg  & 0.05 \cotwo kg &   35 \\
    \rule{0pt}{2ex}%
    train$^{*}$ & time         &  39.22 days      & 5.60 days      &   7 \\
    train$^{*}$ & energy       & 324.75 kWh       & 46.39 kWh      &   7 \\
    train$^{*}$ & emissions    &  68.20 \cotwo kg & 9.74 \cotwo kg &   7 \\
    \bottomrule
    \end{tabular}
  \end{subtable}

  \hfill 

  \begin{subtable}[b]{\textwidth} 
    \caption{All VIT experiment resources.}
    \centering
    \begin{tabular}{llllr}
    \toprule
            Node & Resource &           Total &           $\mu$ &  n \\
    \midrule
    eval        &        time &    5.84 days     &  0.35 hours    &  399 \\
    \rule{0pt}{2ex}%
    pred        &        time &    7.29 days     &  0.44 hours    &  399 \\
    pred        &      energy &  102.83 kWh      &   0.26 kWh     &  399 \\
    pred        &   emissions &  21.6 \cotwo kg  & 0.05 \cotwo kg &  399 \\
    \rule{0pt}{2ex}%
    train$^{*}$ & time        & 158.95 days      &     3.78 days  &   42 \\
    train$^{*}$ & energy      & 1,316.07 kWh     &     31.34 kWh  &   42 \\
    train$^{*}$ & emissions   & 276.37 \cotwo kg & 6.58 \cotwo kg &   42 \\
    \bottomrule
    \end{tabular}
  \end{subtable}
\ifwacv \VitResourceCaption \fi
\end{table*}

\subsubsection{VIT Resource Usage}
\label{sec:vit_environmental_impact}

Note:
we remind the reader that this section only applies to the VIT models.

All models were trained on a single machine with an 11900k CPU and a 3090 GPU.
At predict time, using one background worker, our models processed 416 $\times$ 416 patches at a rate of
  20.93Hz with 94\% GPU utilization.

To better understand the energy requirements of our model, particularly for potential deployment on mobile
  devices, we used CodeCarbon \cite{lacoste2019codecarbon} to measure the resource usage during prediction and
  evaluation.
This analysis not only informs practical considerations but also helps us assess our contribution to the
  growing carbon footprint of AI \cite{kirkpatrick_carbon_2023}.
The results for the 7 presented training experiments and the total 42 training experiments are reported in
  \Cref{tab:resources}.

Direct measurement of resource usage during training is still under development, but we estimate the
  duration of each training run using indirect methods.
We approximate energy consumption by assuming a constant power draw of 345W from the 3090 GPU during
  training.
Emissions are estimated using a conversion ratio of 0.21 $\frac{\textrm{kg}\cotwo{}}{\textrm{kWh}}$.
  
Based on the validation set's 691 images, we estimate that predicting on a single image on our desktop
  requires approximately 1.15 seconds and 0.13 Wh of energy.
For context, typical mobile phones have a battery capacity of around 10 Wh and significantly less compute
  power than our desktop setup.
While our models demonstrate the feasibility of training a strong detector from our dataset, they are not
  optimized for the mobile setting.
To deploy our model on mobile devices, we will need to improve its efficiency or explore more efficient
  architectures.

\section{Environmental Impact} 
\label{sec:general_environmental_impact}

A footnote in the main paper reports the experiment costs, here we expand on the details.
These costs are the total over all runs in the development of this paper over different datasets, with
  different numbers of runs per model, so it cannot be used to infer running time of the models.
Instead it reports a component of the cost of performing this research.
All costs are estimated assuming \$0.16 per kWh, \$25 per 1000~kg CO$_2$.
The breakdown of resources used is given in \Cref{tab:resources_breakdown}.

Training accounted for the majority of resource usage with VIT models being the most expensive to run
  (\Cref{sec:vit_environmental_impact}).
The main reason is that VIT experiments operated on half-resolution images (2,016~$\times$~1,512) using
  416~$\times$~416 patches, whereas GroundingDINO, YOLO, and MaskRCNN were trained on smaller resized inputs
  (e.g., 640~$\times$~640, 1066~$\times$~800, depending on framework defaults) without windowing.
A second reason is that we trained many more VIT variants in a hyperparameter search, which was done before
  easy to use foundational models became available.
For training we estimated energy usage by measuring time and estimating GPU power draw approximated at 345W.
To estimate emissions we used a factor of 0.21~kg CO$_2$/kWh.

For prediction resources estimates, there is an important limitation.
The system running experiments was equipped with two RTX~3090 GPUs, although only a single one was used for
  any individual experiments.
However, due to our use of CodeCarbon, which counts entire system resources, some double counting may have
  occurred when we ran two experiments simultaneously.
Not all experiments were run in parallel, but some were.
Still our estimates provide an upper bound for the resources utilization and we the lower bound will at best
  be half of our reported numbers.
This limitation does not apply to our training estimations, which is the bulk of our cost, and thus our
  total numbers should only slightly inflated.

\newcommand{\TotalResourceCaption}{
\caption{Resource usage for training and prediction by model family.
Time is wall-clock duration on a single RTX~3090.
Energy is electricity consumed.
Emissions use a 0.21~\cotwo kg/kWh factor.
Cost is estimated at \$0.16/kWh electricity and \$25 per 1000~\cotwo kg.}
\label{tab:resources_breakdown}
}
\begin{table*}[hb]
\ifwacv \else \TotalResourceCaption \fi
\centering

\begin{tabular}{llrrrr}
\toprule
Phase & Model family   & Time (days) & Energy (kWh) & Emissions (\cotwo kg) & Cost (USD) \\
\midrule
Train
 & VIT-sseg      & 158.95 & 1316.07 & 276.37 & 217.48 \\
 & MaskRCNN      & 0.71   & 5.92    & 1.24   & 0.98 \\
 & YOLO-v9       & 4.14   & 34.30   & 7.20   & 5.67 \\
 & Grounding DINO& 0.32   & 2.68    & 0.56   & 0.44 \\
 & \textbf{Total (training)} & \textbf{164.12} & \textbf{1358.96} & \textbf{285.38} & \textbf{224.57} \\
\midrule
Test
 & VIT-sseg      & 13.13  & 102.83  & 21.60   & 16.99 \\
 & MaskRCNN      & 0.57   & 4.41    & 0.93    & 0.73 \\
 & YOLO-v9       & 0.08   & 0.19    & 0.02    & 0.03 \\
 & Grounding DINO& 0.13   & 0.29    & 0.02    & 0.05 \\
 & \textbf{Total (prediction)} & \textbf{13.91} & \textbf{107.72} & \textbf{22.57} & \textbf{17.80} \\
\midrule
\multicolumn{2}{l}{\textbf{Overall total}} & \textbf{178.03} & \textbf{1466.69} & \textbf{307.95} & \textbf{242.37} \\
\bottomrule
\end{tabular}
\ifwacv \TotalResourceCaption \fi
\end{table*}

\section{Changes Since arXiv v1 (2412.16473v1)}
\label{sec:changes-since-v1}

This appendix summarizes changes relative to the original preprint
\redact{(\url{https://arxiv.org/abs/2412.16473v1})}.

\begin{itemize}
    \setlength{\itemsep}{0pt}
    \setlength{\parskip}{0pt}
    \setlength{\parsep}{0pt}

    \item \textbf{Dataset size:} We expanded the dataset from 42GB of \textasciitilde6k full-resolution images with \textasciitilde4k polygon annotations (v1) to 60GB of $>$9k images with \textasciitilde6k polygon annotations (this version).

    \item \textbf{Evaluation:} We expand the independently collected, community-contributed test set from 30 images (v1) to 121 images (this version).

    \item \textbf{Baselines and experiments:} In addition to ViT and Mask R-CNN baselines in v1, we include YOLO-v9 and DINO-v2 baselines in this version, and we also report zero-shot results with GroundingDINO. We additionally distinguish scratch vs.\ pretrained training for the trainable baseline families where applicable.

    \item \textbf{Distribution experiment:} The distribution experiment in this version still corresponds to the v1 dataset (i.e., the distribution study was not re-run on the expanded dataset).

    \item \textbf{Paper revisions:} We clarify parts of the methodology (e.g., BAN) and update figures, tables, and text for the expanded dataset and experiments.

\end{itemize}

This work went through three rounds of peer review; we thank the reviewers for
their time and comments, which informed several of the updates above.  This
version corresponds to our accepted paper at the \emph{International Workshop
on Smart Waste Monitoring (WasteVision)} at WACV.

\fi

\end{document}